\documentclass{article}

\usepackage[preprint]{neurips_2026}
\usepackage{xspace}
\newcommand{\eg}{e.g.\@\xspace}

\usepackage[utf8]{inputenc} 
\usepackage[T1]{fontenc}    
\usepackage{hyperref}       
\usepackage{url}            
\usepackage{booktabs}       
\usepackage{amsfonts}       
\usepackage{nicefrac}       
\usepackage{microtype}      
\usepackage{xcolor}         
\usepackage{graphicx}      
\usepackage{eso-pic} 
\usepackage{amsmath}
\usepackage{natbib}
\usepackage{multirow}
\usepackage[table]{xcolor}
\usepackage{enumitem}
\usepackage[dvipsnames]{xcolor} 
\usepackage{xcolor}
\usepackage{threeparttable}
\usepackage{caption} 
\usepackage{graphicx}
\usepackage{wrapfig}
\usepackage{arydshln}
\usepackage{wrapfig}
\usepackage{booktabs}
\usepackage{algorithm}
\usepackage{algorithmic}
\usepackage[most]{tcolorbox}
\usepackage{tikz}
\usepackage{amssymb}
\newcommand{\cmark}{\textcolor{green!60!black}{\checkmark}}
\newcommand{\xmark}{\textcolor{red!70!black}{$\times$}}

\definecolor{boxbackground}{HTML}{F0F7FF}
\definecolor{boxborder}{HTML}{D0D9E5}
\definecolor{accentblue}{HTML}{4A86E8}
\newtcolorbox{promptbox}[2][]{%
    enhanced,
    breakable,
    boxsep=5pt,
    left=9pt,
    right=7pt,
    top=5pt,
    bottom=5pt,
    colback=boxbackground,
    colframe=boxborder,
    boxrule=0.5pt,
    arc=4pt,
    frame hidden,
    borderline west={3pt}{0pt}{accentblue},
    shadow={0.5pt}{0.5pt}{1.5pt}{black!10},
    fontupper=\normalsize,
    title=#2,
    colbacktitle=accentblue,
    coltitle=white,
    fonttitle=\fontsize{9}{11}\selectfont\bfseries,
    attach boxed title to top left={yshift=-2.5mm, xshift=3.2mm},
    boxed title style={
        enhanced,
        left=3pt,
        right=3pt,
        top=1pt,
        bottom=1pt,
        boxsep=2pt,
        arc=3pt,
        boxrule=0pt,
        colback=accentblue,
    },
    #1
}

    \usepackage{xfp} 
    \usepackage{pgf}
    \AddToShipoutPictureBG{%
      \ifnum\value{page}=1
        \AtPageUpperLeft{%
          \put(\LenToUnit{3.8cm},\LenToUnit{-2.7cm}){%
            \includegraphics[height=0.8cm]{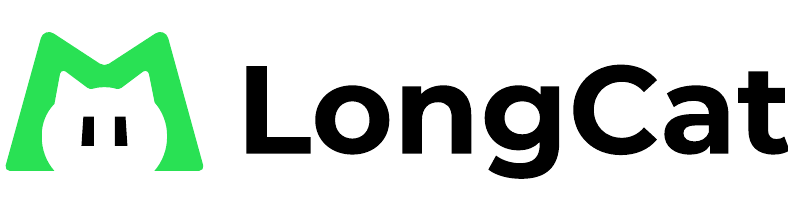}%
          }%
        }%
      \fi
    }

\title{Look Before You Leap: Autonomous Exploration for LLM Agents}

\author{%
  \textbf{Ziang Ye\textsuperscript{1,2}\thanks{Work was done during the internship in Meituan.}}\quad
\textbf{Wentao Shi\textsuperscript{1}}\quad
\textbf{Yuxin Liu\textsuperscript{1,2}}\quad
\textbf{Yu Wang\textsuperscript{1,2}}\quad
\textbf{Zhengzhou Cai\textsuperscript{1,2}}\quad
\textbf{Yaorui Shi\textsuperscript{1,2}}\quad \\
\textbf{Qi Gu\textsuperscript{2}}\thanks{Corresponding Author}\quad 
\textbf{Xunliang Cai\textsuperscript{2}}\quad 
\textbf{Fuli Feng\textsuperscript{1}}\footnotemark[\value{footnote}]\quad 
\\
\textsuperscript{1}University of Science and Technology of China \quad
\textsuperscript{2}Meituan \quad 
\\
\small{
  \texttt{yza03@mail.ustc.edu.cn} \quad \texttt{\small guqi03@meituan.com}
} 
\small{
\texttt{fulifeng93@gmail.com} 
}
}

\begin{document}

\maketitle
\vspace{-16pt}

\begin{abstract}
Large language model based agents often fail in unfamiliar environments due to premature exploitation: a tendency to act on prior knowledge before acquiring sufficient environment-specific information. We identify autonomous exploration as a critical yet underexplored capability for building adaptive agents. To formalize and quantify this capability, we introduce Exploration Checkpoint Coverage, a verifiable metric that measures how broadly an agent discovers key states, objects, and affordances. Our systematic evaluation reveals that agents trained with standard task-oriented reinforcement learning consistently exhibit narrow and repetitive behaviors that impede downstream performance. To address this limitation, we develop a training strategy that interleaves task-execution rollouts and exploration rollouts, with each type of rollout optimized
by its corresponding verifiable reward. Building on this training strategy, we propose the Explore-then-Act paradigm, which decouples information-gathering from task execution: agents first utilize an interaction budget to acquire grounded environmental knowledge, then leverage it for task resolution. Our results demonstrate that learning to systematically explore is imperative for building generalizable and real-world-ready agents.
\end{abstract} 

\section{Introduction}
\label{sec:intro}
Large language model based agents have remarkable application in realistic scenarios involving multi-step interactions with complex and diverse environments~\cite{agentbench,WebArena,osworld,tau2,swebench}. With the advancement of Reinforcement Learning with Verifiable Rewards (RLVR), models have made substantial progress in interacting with complex environments to solve multi-step tasks~\citep{ragen,agentgym-rl,gigpo}.
Despite this progress, a key aspect remains underexplored: current RLVR approaches primarily optimize for task-completion rewards in known or static distributions, thereby encouraging instrumental behaviors aimed at solving predefined tasks. As a result, they provide limited incentive for developing the autonomous exploration capabilities required to adapt to novel, unfamiliar environments.

In the absence of intrinsic exploratory capability, current LLM-based agents often exhibit a pattern of \textbf{premature exploitation}. When deployed in an unfamiliar environment, these agents tend to prematurely commit to actions derived from training-time priors, rather than systematically interacting with their surroundings to uncover hidden constraints or identify available tools~\citep{walle,tttadapt}. This limitation manifests in two recurring failure modes. First, the agent often lacks a clear starting point. As a result, it either engages in aimless trial and error or confidently follows a poorly informed plan, rather than proactively acquiring task-relevant state information~\citep{fundamentals,Agent-R}. Second, the agent might misinterpret environment-specific semantics, such as specific tool arguments or UI affordances, leading to action-environment mismatches that accumulate into failures~\citep{jiang2025verltool,mcpatlas}.

To alleviate the inadequate environment understanding problem, prior work has primarily focused on preparing environment-specific knowledge before deployment. 
Several methods construct diverse task sets that broadly cover target environments, encouraging models to internalize environment-specific knowledge during training~\citep{cues,learnbyinteract,explorer}; 
Others build external knowledge bases or manuals through complex frameworks that model the environment~\citep{walle,wese,automanual}. 
Although these approaches can improve performance in their target environments, they rely on pre-compiling knowledge offline into model weights or external databases, leaving agents without the ability to autonomously acquire environment knowledge online.
This limitation becomes increasingly critical as real-world deployment environments span diverse and dynamically evolving scenarios~\citep{vitabench,envscaler,browsecomp}, where it is infeasible to pre-compile all necessary knowledge. 
This motivates a shift from pre-deploying environment knowledge to endowing agents with the ability to acquire such knowledge themselves through autonomous online exploration.

In this work, we begin by formalizing environment exploration as an independent,
measurable capability and introduce Exploration Checkpoint Coverage (ECC), a
verifiable metric that quantifies the extent to which an agent discovers key states, objects, and affordances in an unfamiliar environment. Using ECC, we conduct a
systematic evaluation of existing models and training paradigms, revealing a
notable finding: task-oriented training, including strong RLVR-style optimization
for task completion, does not reliably yield autonomous exploration ability.
Agents trained under these paradigms often terminate exploration prematurely,
covers only a limited portion of the environment, or interacts repeatedly with a
narrow set of familiar states.

Motivated by this gap, we study how to equip agents with exploration capabilities by explicitly optimizing exploration during training. To achieve this, we introduce an interleaved GRPO training strategy that interleaves task-execution rollouts and exploration rollouts, with each type of rollout optimized by its corresponding verifiable reward. Task-execution rollouts are trained with task-completion rewards, whereas exploration rollouts are trained with the ECC reward to encourage broad coverage of informative states, relevant objects, and available affordances. Building on this training strategy, we introduce the \emph{Explore-then-Act} paradigm: an exploration-capable agent first allocates an interaction budget to autonomously acquire grounded knowledge about the environment and then uses this knowledge to solve the specific task.

We conduct experiments across three diverse interactive environments: ALFWorld~\citep{alfworld}, SciWorld~\citep{scienceworld}, TextCraft~\citep{agentgym}, and a challenging ALFWorld variant. 
Our results show that a wide range of open-source models and task-oriented training paradigms fail to reliably produce meaningful exploration. In contrast, explicitly training agents to explore develops this capability and substantially improves downstream task performance. Moreover, exploration-aware models can more effectively convert an initial interaction budget into useful environment knowledge, leading to stronger downstream task performance. These results suggest that autonomous exploration serves as a key meta-capability that enables agents to acquire grounded environment knowledge before acting, thereby improving adaptability and generalization in unfamiliar environments.

Our contributions can be summarized as follows:
\begin{itemize}[leftmargin=*, parsep=1pt]
\item We formalize autonomous environment exploration as an independent agent capability and introduce Exploration Checkpoint Coverage (ECC), a verifiable metric for measuring exploration coverage.

\item We systematically demonstrate that task-oriented training, fails to reliably yield autonomous exploration. To address this limitation, we develop an effective training strategy that optimizes for exploration capabilities through interleaved GRPO with an ECC reward.

\item We propose Explore-then-Act, a paradigm that lets agents acquire environment knowledge before task execution, leading to improved downstream performance and robustness across diverse environments and challenging variants.

\item We provide extensive experiments demonstrating that our ECC-guided exploration training substantially improves exploration coverage, downstream task performance, and robustness over task-oriented training baselines.
\end{itemize}

\vspace{-1.2em}
\section{Related Work}
\vspace{-0.6em}
\subsection{LLM-based Agents}

Large language models (LLMs) have become foundational components in modern agent systems, owing to their strong instruction-following capabilities, robust planning abilities, and broad generalization across diverse environments~\cite{voyager,realworld,osatlas,codeagent}. The development of LLM-based agents has evolved through several paradigms. Initial approaches primarily utilized prompt engineering~\citep{react_paradigm,reflexion_paradigm,automanual}, whereas subsequent methods enhanced agent performance through supervised fine-tuning on curated trajectories~\citep{agenttuning,toolllm,gorilla,catastrophicforgetting}. Nevertheless, these methods are often constrained by the narrow scope of their training data, which limits their generalization to novel settings. More recently, reinforcement learning has emerged as a promising alternative~\citep{agentrl,agentgym-rl,gigpo,hgpo}, wherein agents are optimized via policy-gradient methods based on task-completion rewards. Across all these paradigms, however, a common limitation is that agents are typically optimized solely for task reward, lacking an explicit incentive for the information-gathering behavior required in unfamiliar environments. Consequently, they remain susceptible to premature exploitation when subjected to distributional shifts.

\vspace{-0.5em}

\subsection{Environment Modeling for Agents}
To bridge the discrepancy between the training-time priors of LLM-based agents and the dynamics of unfamiliar environments, existing literature has predominantly formulated environment modeling as an offline engineering or pre-compilation task. One prominent line of research employs heuristic or code-driven pipelines to construct external knowledge bases. For instance, frameworks such as Wall-E~\cite{walle}, WESE~\cite{wese}, and AutoManual~\cite{automanual} typically rely on traditional search algorithms (e.g., BFS or DFS) or extensive hand-crafted scripts to systematically probe the environment, utilizing the LLM exclusively to parse observations into structured graphs or rules. An alternative trajectory, exemplified by CUES~\cite{cues}, Learn-by-Interact~\cite{learnbyinteract}, and Explorer~\cite{explorer}, attempts to instill environment knowledge by substantially expanding the diversity of training tasks. This approach effectively compels the model to internalize the constraints of specific environments during the training phase. Nevertheless, all such paradigms fundamentally remain tethered to static, offline mechanisms rather than cultivating the intrinsic, online exploration capabilities necessary for true autonomous adaptability.

\begin{figure*}[t]
    \centering
    \includegraphics[width=0.95\textwidth]{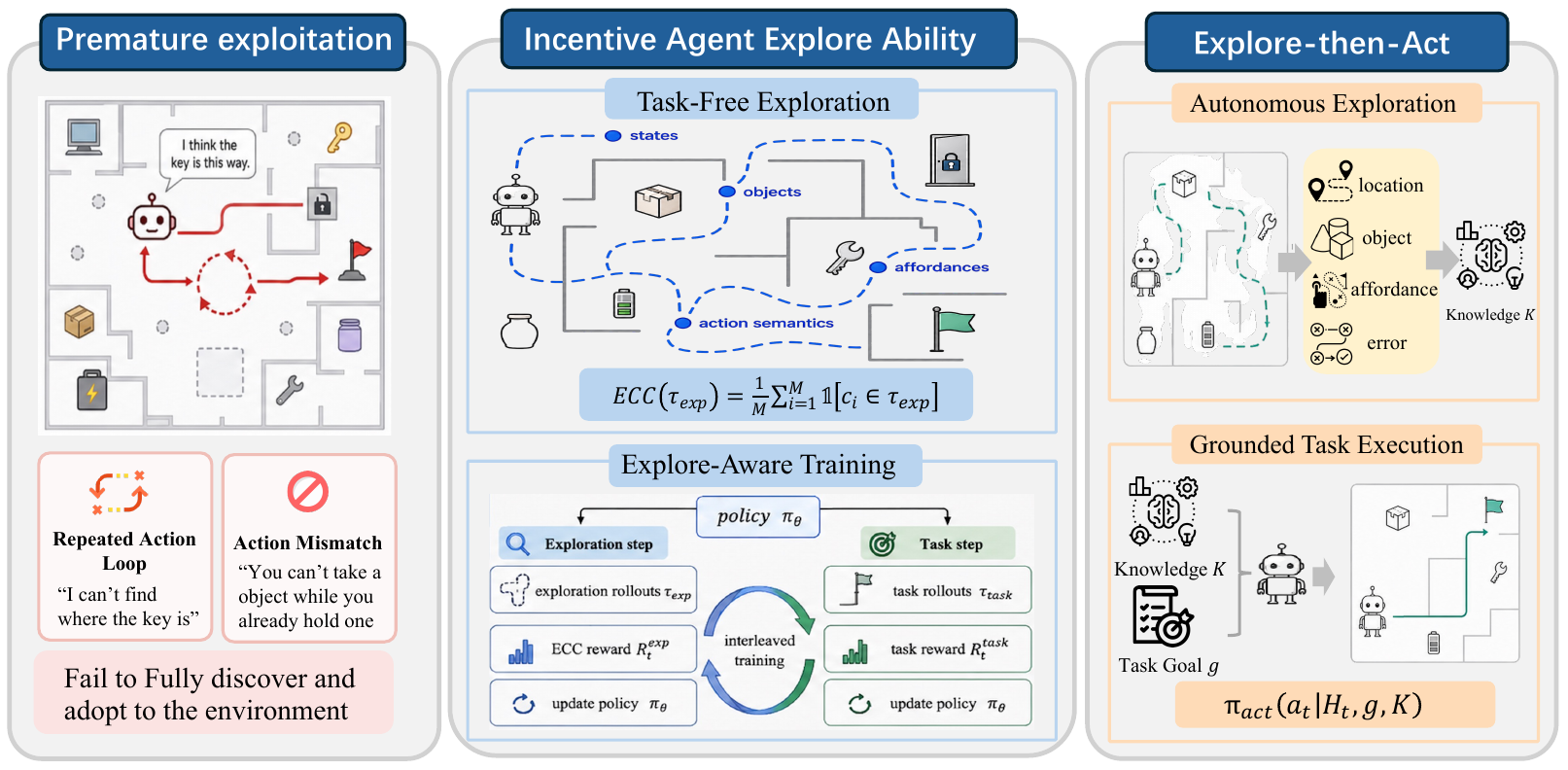}\
    \caption{
Task-oriented training fails to produce autonomous exploration capabilities, resulting in agents that prematurely exploit familiar patterns and acquire limited environment knowledge. We explicitly optimize for exploration through ECC rewards, enabling agents to systematically discover environment structure, objects, and affordances. The resulting Explore-then-Act paradigm decouples information gathering from task execution: agents first explore to acquire grounded knowledge, then leverage it to solve downstream tasks.
}
    \label{fig:main}
\end{figure*}

\vspace{-0.5em}
\section{Methodology}
\label{sec:method}
In this work, we investigate autonomous environment exploration as an independent capability of LLM-based agents. Rather than treating exploration as a mere byproduct of task execution, we formalize it as a goal-free, information-gathering process wherein an agent actively probes an unfamiliar environment to uncover intrinsic states, available objects, functional affordances, and action semantics. To rigorously quantify this behavior, we introduce Exploration Checkpoint Coverage (ECC) as a verifiable metric of exploration quality, and we examine methodologies to explicitly optimize agents for this capability. Finally, we demonstrate how the knowledge acquired through autonomous exploration can be systematically leveraged to enhance downstream task execution via an Explore-then-Act protocol. As illustrated in Figure~\ref{fig:main}, our framework addresses the limitation of task-oriented training, which tends to induce premature exploitation, by explicitly rewarding broad environment discovery and separating the exploration phase from the subsequent task goal-conditioned acting phase.

\vspace{-0.3em}
\subsection{Problem Formulation}
\label{subsec:method:formulation}

We begin by formalizing the standard task setting for agents and subsequently define autonomous exploration as a distinct interaction process.

\vspace{-0.3em}
\subsubsection{Agent environmment Interaction}
\label{subsubsec:method:preliminaries}
We consider a standard setting where an LLM-based agent interacts with an environment $\mathcal{E}$. The agent's objective is to complete a task specified by a high-level natural language goal, $g$. The interaction unfolds over a sequence of steps. At each step $t$, the agent receives an observation $o_t \in \mathcal{O}$ from the environment, which describes the current state. Based on the history of interactions $H_t = (o_1, a_1, \dots, o_t)$, the agent's policy $\pi$ generates the next action $a_t \in \mathcal{A}$. The policy is typically conditioned on both the history and the goal: $a_t \sim \pi(\cdot | H_t, g)$.

This multi-step interaction produces a trajectory $\tau = (o_1, a_1, o_2, a_2, \dots, o_T)$, where $T$ is the episode length. The agent's performance is evaluated by a reward function $R(\tau, g) \in \{0, 1\}$, which assigns a reward of 1 upon task success and 0 otherwise. In this conventional paradigm, the agent follows an exploitative behavioral pattern, with each action instrumentally directed toward maximizing the task-specific reward $R$.

\subsubsection{Autonomous Environment Exploration}
\label{subsubsec:method:exploration}
In contrast to goal-directed task execution, we define \textit{autonomous exploration} as a proactive, information-gathering process that operates independently of any specific task goal. In this mode, the agent is situated within the environment $\mathcal{E}$ without an assigned task $g$. Its primary objective shifts to interactively probing the surroundings to build an internal model of the environment's latent transition dynamics $\mathcal{T}(o_{t+1}|o_t, a_t)$, state space (\eg, map layout, available items), and action semantics (\eg, tool arguments, hidden constraints). We formalize this process as an exploration session, which yields a trajectory $\tau_{\text{exp}} = (o_1, a_1, \dots, a_{N}, o_{N+1})$, where $N$ denotes the allocated interaction budget. Subsequently, the agent processes $\tau_{\text{exp}}$ to synthesize a grounded knowledge summary, denoted as $\mathcal{K}$. This knowledge encapsulates the discovered environment-specific characteristics, serving to reconcile the discrepancies between the pre-existing priors of the agent and the actual properties of the environment.

\subsection{Measuring Exploration with Exploration Checkpoint Coverage}
\label{subsec:method:ecc}

\begin{wrapfigure}{r}{0.45\linewidth}
    \vspace{-1em} 
    \centering
    \includegraphics[
        width=0.95\linewidth,
    ]{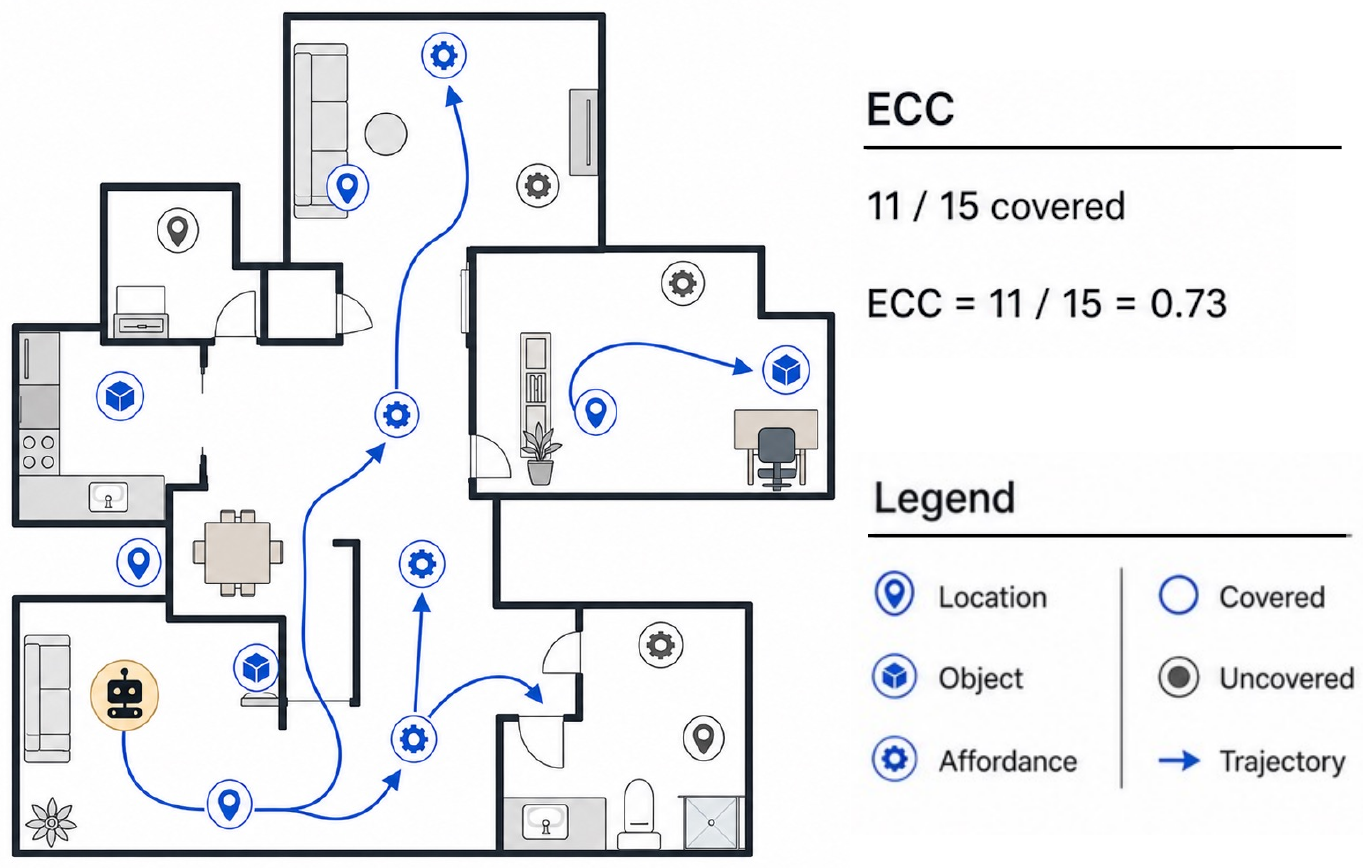}
    \caption{
    Illustration of Exploration Checkpoint Coverage (ECC).
    }
    \label{fig:ecc}
\end{wrapfigure}
To quantify autonomous exploration independently from task success, we introduce \emph{Exploration Checkpoint Coverage} (ECC). For each environment instance, we define a finite set of exploration checkpoints
\begin{equation}
\mathcal{C} = \{c_1, c_2, \dots, c_M\}.
\end{equation}

Each checkpoint corresponds to an environment-specific fact or affordance that a competent explorer should be able to discover. Examples include reachable locations, important objects, valid interaction targets, functional states, action-relevant affordances, or environment-specific constraints.

Given an exploration trajectory $\tau_{\textsc{exp}}$, we define a binary indicator
$\mathbb{1}[c_i \in \tau_{\textsc{exp}}]$ that equals 1 if checkpoint $c_i$ is reached, observed, or otherwise verified during exploration. ECC is computed as the fraction of checkpoints covered:

\begin{equation}
\textsc{ECC}(\tau_{\textsc{exp}}) =
\frac{1}{M} \sum_{i=1}^{M} \mathbb{1}[c_i \in \tau_{\textsc{exp}}].
\end{equation}

We provide an intuitive illustration in Figure~\ref{fig:ecc} to demonstrate environment checkpoints and ECC calculation. Details of checkpoint generation are provided in Appendix~\ref{sec:appendix:ECCcondtruction}.

\subsection{Training Exploration-Capable Agents}
\label{subsec:method:training}

Having formalized autonomous exploration as a measurable capability, we now detail how to explicitly optimize for it during training. We adapt the Group Relative Policy Optimization (GRPO) framework to directly reward exploration and integrate this process into an interleaved training schedule alongside standard task-oriented optimization.

\paragraph{Optimizing for Exploration via GRPO.}
Our core strategy is to provide a direct learning signal for exploration. For an exploration-focused training step, we define the reward for a rollout $\tau_{\textsc{exp}}$ as its Exploration Checkpoint Coverage:
\begin{equation}
R_{\textsc{exp}}(\tau_{\textsc{exp}}) = \textsc{ECC}(\tau_{\textsc{exp}}).
\end{equation}
This reward directly encourages the agent to discover more environment checkpoints. Because ECC is computed from verifiable environment interactions, this reward signal does not require a subjective, open-ended language judge.

To update the policy, we follow the GRPO procedure. For each exploration context $x$, which consists of an environment instance and a general exploration instruction, we sample a group of $G$ rollouts $\{y^{(i)}\}_{i=1}^{G}$ from the current policy $\pi_{\theta}$. We then compute the ECC reward $R^{(i)} = \textsc{ECC}(\tau_{\textsc{exp}}^{(i)})$ for each rollout and normalize these rewards within the group to obtain relative advantages:
\begin{equation}
A^{(i)} = \frac{R^{(i)} - \mathrm{mean}_{j}(R^{(j)})}{\mathrm{std}_{j}(R^{(j)}) + \epsilon}.
\end{equation}
The policy is then updated to increase the likelihood of trajectories with higher relative ECC, regularized by a KL penalty to maintain stability with respect to a reference model:
\begin{equation}
\max_{\theta}\;
\mathbb{E}_{x}
\left[
\frac{1}{G}
\sum_{i=1}^{G}
A^{(i)}
\log \pi_{\theta}(y^{(i)} \mid x)
-
\beta
\textsc{KL}
\left(
\pi_{\theta}(\cdot \mid x)
\,\|\, 
\pi_{\textsc{ref}}(\cdot \mid x)
\right)
\right].
\end{equation}

\paragraph{Interleaved Training Schedule.}
To develop both exploration and task-solving abilities, we employ an interleaved training schedule that alternates between exploration-focused and task-focused optimization steps. In an exploration step, we update the policy using the ECC-based GRPO objective described above. In a task-execution step, we revert to the standard GRPO setup, where rollouts are generated for specific downstream tasks and rewarded based on task completion. By alternating between these two objectives, our training process enables the agent to cultivate a robust exploration capability while simultaneously learning to apply the acquired knowledge to solve specific goals. The exploration reward provides explicit supervision for discovering environment structure, while the task reward ensures that this capability is effectively leveraged for downstream performance.

\subsection{Explore-then-Act: Decoupling Information Gathering from Task Execution}
\label{subsec:method:eta}
Existing LLM agents predominantly operate under a \emph{direct task-execution} paradigm~\citep{react_paradigm,reflexion_paradigm}, wherein every interaction is strictly conditioned on a specified goal $g$ and evaluated solely by extrinsic task rewards. A canonical instantiation of this approach is the ReAct-style loop, which interleaves reasoning and actions under a unified goal-directed policy, formalized as 
\begin{equation} a_t 
\sim \pi_{\textsc{act}}(\cdot \mid H_t, g), 
\end{equation} 
thereby lacking an explicit mechanism to allocate an interaction budget for resolving environmental uncertainties. To address this limitation, we propose \emph{Explore-then-Act}, an alternative inference paradigm that explicitly decouples environment understanding from goal completion by introducing a preliminary, goal-free exploration phase. During this initial stage, the agent is deployed in the environment $\mathcal{E}$ without a designated task. It follows an exploration policy for a fixed interaction budget of $N$ steps, generating a trajectory $\tau_{\textsc{exp}} = (o_1, a_1, \dots, o_{N}, a_{N}, o_{N+1})$, where 
\begin{equation} 
a_t \sim \pi_{\textsc{exp}}(\cdot \mid H_t). 
\end{equation} 
After completing exploration, the agent synthesizes the interaction sequence into a grounded knowledge summary
$ \mathcal{K} = \textsc{Summarize}(\tau_{\textsc{exp}})$
which serves as a structured natural-language artifact capturing actionable properties of the environment, including state layouts, object affordances, action preconditions, discovered constraints, and failure cases. In the subsequent goal-conditioned acting stage, the agent tackles the downstream task using an updated policy that conditions on the current interaction history, the task goal, and the acquired knowledge:
\begin{equation} a_t \sim \pi_{\textsc{act}}(\cdot \mid H_t, g, \mathcal{K}). \end{equation} 
In practice, this decoupling is implemented by injecting the synthesized knowledge $\mathcal{K}$ into the prompt after the agent completes exploration, ensuring that downstream decisions are grounded in empirically discovered facts about the environment.

\vspace{-0.3em}
\section{Experiments}
\label{sec:experiments}

In this section, we provide a comprehensive evaluation of our proposed framework. We begin by detailing the experimental setup, then examine the inherent exploration deficiencies of contemporary large language models, and finally show that explicit exploration-aware training improves agents’ task-execution capabilities while further transforming the \emph{Explore-then-Act} (E-t-A) paradigm into consistent performance gains.

\subsection{Experimental Setup}

\paragraph{LLM Backbones.}
To ensure the robustness of our conclusions across different model scales and families, we evaluate a diverse set of open-source backbones, including Qwen2.5-7B~\citep{qwen2.5}, Qwen3-4B~\citep{qwen3}, and LLaMA3.1-8B~\citep{llama1}. we also benchmark frontier proprietary models, including GPT-4.1~\citep{GPT4} and Claude-Opus-4.5~\citep{claudeopus}.
\paragraph{Environments.}
We evaluate our approach across three diverse environments, each requiring agents to acquire environment-specific knowledge for effective decision-making. ALFWorld~\citep{alfworld} involves household navigation and object manipulation under high-level instructions. ScienceWorld~\citep{scienceworld} requires agents to discover and apply scientific rules through interactions with a complex simulated world. TextCraft~\citep{agentgym} tests resource gathering and multi-step crafting under hidden recipe structures. Together, these environments cover embodied navigation, scientific reasoning, and compositional planning, providing a comprehensive testbed for exploration and task execution.

\subsection{Diagnosing the Exploration Deficit in Current LLMs Agents}
\label{subsec:exp_diagnosis}

Before evaluating downstream task completion, we must answer a fundamental question: \emph{How thoroughly can current LLMs autonomously discover their environment without explicit task guidance?}

\paragraph{Implementation Details.}
We deploy each LLM Agent in all three environments, imposing a maximum interaction budget of 100 steps. Crucially, the agents are \textbf{not} provided with any specific task instructions. Instead, they are prompted to freely explore and interact with the environment to gather as much useful information as possible. Detailed specifications of the prompts, and the ECC construction details are provided in Appendix~\ref{sec:appendix:prompt} and Appendix~\ref{sec:appendix:ECCcondtruction}, respectively.

\paragraph{Evaluation Metrics.}
We evaluate exploration quality using two metrics: average trajectory length (\textbf{Steps}) and Exploration Checkpoint Coverage (\textbf{ECC}, \%), as defined in Section~\ref{subsec:method:ecc}. ECC quantifies the fraction of predefined environment checkpoints discovered during the free exploration phase, including critical states, key objects, and distinct locations. We then measure the downstream utility of exploration by reporting the \textbf{Performance Gain}, denoted as $\Delta_{\text{Task}}=\text{E-t-A}-\text{Dir.}$, which captures the absolute improvement in task success rate under the Explore-then-Act paradigm (E-t-A) over direct task execution without prior exploration (Dir.). To ensure a fair comparison, all downstream task executions are performed by a fixed agent, Qwen3-4B, thereby isolating the exploration model as the only varying component.


\newcommand{\rlTag}{\textsubscript{\textsc{+GRPO}}}

\begin{table*}[!t]
\centering
\caption{
\textbf{Autonomous exploration capability in task-free environments.}
We place each agent in three environments without task instructions and ask it to
freely explore within a budget of 100 steps.
We report average interaction turns (\textbf{Steps}), Exploration Checkpoint
Coverage (\textbf{ECC}, \%), and the downstream task performance change induced
by Explore-then-Act, denoted as
$\Delta_{\text{Task}}=\text{E-t-A}-\text{Dir.}$.
The rightmost columns report the macro-average ECC and $\Delta_{\text{Task}}$
across all three environments.
}
\label{tab:exploration_deficit}
\vspace{4pt}

\setlength{\tabcolsep}{4pt}
\renewcommand{\arraystretch}{1.2}
\resizebox{0.95\textwidth}{!}{%
\begin{tabular}{@{} l ccc ccc ccc cc @{}}
\toprule

\multirow{2}{*}{\textbf{Model}}
& \multicolumn{3}{c}{\textbf{ALFWorld}}
& \multicolumn{3}{c}{\textbf{SciWorld}}
& \multicolumn{3}{c}{\textbf{TextCraft}}
& \multirow{2}{*}{\textbf{Avg.\ ECC}}
& \multirow{2}{*}{$\boldsymbol{\Delta_{\text{Task}}}$} \\
\cmidrule(lr){2-4}
\cmidrule(lr){5-7}
\cmidrule(lr){8-10}
& \small{Steps} & \small{ECC} & \small{$\Delta$}
& \small{Steps} & \small{ECC} & \small{$\Delta$}
& \small{Steps} & \small{ECC} & \small{$\Delta$}
& & \\

\midrule

\rowcolor{gray!10}
\multicolumn{12}{l}{\textbf{\textit{Open-Source Models}}} \\[1pt]

Qwen2.5-7B
  & 36.8 & 19.3 & $-0.3$
  & 63.4 & 32.1 & $-0.6$
  & 50.8 & 15.2 & $-1.1$
  & 22.2 & $-0.7$ \\

Qwen2.5-7B\rlTag
  & 11.8 & 11.2 & $-1.3$
  & 7.4  & 15.4 & $-0.3$
  & 8.7  & 11.3 & $-2.1$
  & 12.6 & $-1.2$ \\[3pt]

Qwen3-4B
  & 19.2 & 35.5 & $-2.2$
  & 87.8 & 29.3 & $-0.9$
  & 21.9 & 20.6 & $-3.4$
  & 28.5 & $-2.2$ \\

Qwen3-4B\rlTag
  & 35.5 & 32.8 & $-0.5$
  & 43.4 & 12.9 & $-1.7$
  & 14.5 & 10.8 & $-0.2$
  & 18.8 & $-0.8$ \\[3pt]

LLaMA3.1-8B
  & 22.5 & 36.8 & $-1.6$
  & 97.5 & 33.7 & $-2.1$
  & 65.9 & 22.1 & $-1.5$
  & 30.9 & $-1.7$ \\

\midrule

\rowcolor{gray!10}
\multicolumn{12}{l}{\textbf{\textit{Closed-Source Models}}} \\[1pt]

GPT-4.1
  & 24.8 & 52.3 & $+1.9$
  & 50.8 & 38.7 & $-0.2$
  & 31.4 & 57.6 & $+4.3$
  & 49.3 & $+2.0$ \\

Claude-Opus-4.5
  & 61.9 & \textbf{96.8} & $+6.3$
  & 97.8 & \textbf{89.3} & $+11.7$
  & 97.3 & \textbf{82.5} & $+7.8$
  & \textbf{89.5} & $+8.6$ \\

\bottomrule
\end{tabular}%
}
\vspace{-6pt}
\end{table*}

\paragraph{Results.}
The results, presented in Table~\ref{tab:exploration_deficit}, reveal a significant exploration deficit in existing models. Our analysis reveals three primary findings:
\begin{itemize}[leftmargin=*, itemsep=0pt,parsep=1pt]
    \item \textbf{Open-source models exhibit limited intrinsic exploratory behavior.} Models such as Qwen2.5-7B and Qwen3-4B achieve low average ECC scores (22.2\% and 28.5\%, respectively), frequently becoming trapped in repetitive loops or terminating their exploration prematurely.
    \item \textbf{Task-oriented reinforcement learning can impede exploratory capabilities.} Fine-tuning these models with task-oriented GRPO \emph{reduces} their exploration coverage, as exemplified by Qwen3-4B, whose average ECC drops from 28.5\% to 18.8\%. This finding suggests that optimizing for task-completion rewards fosters narrow, instrumental policies at the expense of systematic environment mapping. 
    \item \textbf{Ineffective exploration can degrade downstream task performance.} Consequently, the Explore-then-Act paradigm is not universally beneficial. When the exploration phase is shallow, repetitive, or misaligned with the environment's structure, the collected observations constitute noisy or incomplete context rather than actionable guidance.
\end{itemize}

\subsection{Equipping LLM Agents with Exploration Abilities}
\label{subsec:exp_incentive}

Given that optimizing for task-specific rewards is insufficient for fostering exploration, we investigate whether reinforcement learning with explicit exploration-aware objectives can instill autonomous exploratory capabilities.

\paragraph{Implementation Details.}
We train agents with Group Relative Policy Optimization (GRPO) under three configurations aligned with the formulations in Section~\ref{sec:method}, using the training split from AgentGym~\citep{agentgym}. All models are trained for up to 300 steps. GRPO (Task-Only) serves as a conventional goal-directed baseline corresponding to Section~\ref{subsubsec:method:preliminaries}, where the agent is optimized solely on task-specific rollouts. GRPO (Explore-Only), corresponding to Section~\ref{subsubsec:method:exploration}, removes explicit task goals and optimizes the policy with the Exploration Checkpoint Coverage (ECC) reward $R_{\textsc{exp}}$, encouraging purely information-seeking behavior. Our main method follows the interleaved training schedule in Section~\ref{subsec:method:training}, alternating between task-focused and exploration-focused updates so that the agent develops both downstream task-solving ability and autonomous exploration capability. Unless otherwise specified, we use a 5:1 ratio of task-execution to exploration rollouts to balance task proficiency and exploration. We provide sensitivity experiments for this parameters in Appendix~\ref{sec:appendix:sensitive}. Detailed hyperparameters and training configurations are provided in Appendix~\ref{sec:appendix:addexpdetails}.

\begin{table*}[!t]
\centering

\definecolor{gaincolor}{RGB}{34,139,34}
\definecolor{dropcolor}{RGB}{220,20,60}

\caption{
We report task success rates across three interactive environments, comparing models trained with and without exploration-aware objectives. Models are evaluated under two execution paradigms: \emph{Direct Execution} (Dir.) and \emph{Explore-then-Act} (E-t-A). The subscript
$\textcolor{purple}{\tiny \uparrow\Delta}$ /
$\textcolor{blue}{\tiny \downarrow\Delta}$
indicates the performance change of the E-t-A paradigm relative to Direct Execution. The results highlight that exploration-aware models consistently benefit from an initial exploration phase, whereas task-only models often exhibit a performance decline.
}
\label{tab:main_results_side_by_side}
\vspace{1.8pt}

\newcommand{\gain}[1]{\textcolor{purple}{\tiny \,$\uparrow$#1}}
\newcommand{\drop}[1]{\textcolor{blue}{\tiny \,$\downarrow$#1}}

\resizebox{0.96\textwidth}{!}{%
\renewcommand{\arraystretch}{1.05}
\setlength{\tabcolsep}{6pt}
\begin{tabular}{@{} l cc cc cc cc @{}}
\toprule

\multirow{2}{*}{\textbf{Method}} 
& \multicolumn{2}{c}{\textbf{ALFWorld}} 
& \multicolumn{2}{c}{\textbf{SciWorld}} 
& \multicolumn{2}{c}{\textbf{TextCraft}} 
& \multicolumn{2}{c}{\textbf{Avg.}} \\
\cmidrule(lr){2-3} \cmidrule(lr){4-5} \cmidrule(lr){6-7} \cmidrule(lr){8-9}
& \small{Dir.} & \small{E-t-A} 
& \small{Dir.} & \small{E-t-A} 
& \small{Dir.} & \small{E-t-A} 
& \small{Dir.} & \small{E-t-A} \\
\midrule

\rowcolor{gray!12} \multicolumn{9}{l}{\textbf{\textit{Backbone: Qwen2.5-7B}}} \\

Zero-Shot (ReAct)
& 54.4 & 54.1\drop{\textbf{0.3}}
& 4.9  & 4.3\drop{\textbf{0.6}}
& 15.4 & 14.3\drop{\textbf{1.1}}
& 24.9 & 24.2\drop{\textbf{0.7}} \\

\addlinespace[3pt]
\multicolumn{9}{l}{\hspace{4pt}\textit{Task-Oriented Baselines} } \\

GRPO (Task-Only)
& 94.4 & 93.2\drop{\textbf{1.2}}
& 43.9 & 42.6\drop{\textbf{1.3}}
& 66.8 & 66.5\drop{\textbf{0.3}}
& 68.4 & 67.4\drop{\textbf{0.9}} \\

\addlinespace[3pt]
\cdashline{1-9}
\addlinespace[2pt]
\multicolumn{9}{l}{\hspace{4pt}\textit{Exploration-Aware Training} } \\
GRPO (Explore-only)
& 60.2 & 65.3\gain{\textbf{5.1}}
& 10.1 & 14.2\gain{\textbf{4.1}}
& 46.7 & 49.7\gain{\textbf{3.0}}
& 39.0 & 43.1\gain{\textbf{4.1}} \\

\rowcolor{blue!5}
GRPO (Interleaved)
& \textbf{96.9} & \textbf{98.5}\gain{\textbf{1.6}}
& \textbf{45.8} & \textbf{47.2}\gain{\textbf{1.4}}
& \textbf{70.1} & \textbf{73.7}\gain{\textbf{3.6}}
& \textbf{70.9} & \textbf{73.1}\gain{\textbf{2.2}} \\

\midrule

\rowcolor{gray!12} \multicolumn{9}{l}{\textbf{\textit{Backbone: Qwen3-4B}}} \\

Zero-Shot (ReAct)
& 30.9 & 28.7\drop{\textbf{2.2}}
& 5.2  & 4.3\drop{\textbf{0.9}}
& 34.1 & 30.7\drop{\textbf{3.4}}
& 23.4 & 21.2\drop{\textbf{2.2}} \\

\addlinespace[3pt]
\multicolumn{9}{l}{\hspace{4pt}\textit{Task-Oriented Baselines} } \\

GRPO (Task-Only)
& 84.6 & 84.3\drop{\textbf{0.3}}
& 54.9 & 53.1\drop{\textbf{1.8}}
& 82.2 & 83.1\gain{\textbf{0.9}}
& 73.9 & 73.5\drop{\textbf{0.4}} \\

\addlinespace[3pt]
\cdashline{1-9}
\addlinespace[2pt]
\multicolumn{9}{l}{\hspace{4pt}\textit{Exploration-Aware Training} } \\

GRPO (Explore-only)
& 49.7 & 54.4\gain{\textbf{4.7}}
& 26.1 & 28.9\gain{\textbf{2.8}}
& 70.3 & 73.1\gain{\textbf{2.8}}
& 48.7 & 52.1\gain{\textbf{3.4}} \\

\rowcolor{blue!5}
GRPO (Interleaved)
& \textbf{90.5} & \textbf{92.7}\gain{\textbf{2.2}}
& \textbf{55.2} & \textbf{56.9}\gain{\textbf{1.7}}
& \textbf{85.9} & \textbf{89.0}\gain{\textbf{3.1}}
& \textbf{77.2} & \textbf{79.5}\gain{\textbf{2.3}} \\

\bottomrule
\end{tabular}
}
\vspace{-5pt}
\end{table*}


\vspace{-1em}
\paragraph{Results.}
Table~\ref{tab:main_results_side_by_side} summarize the task success rates comparing Direct Execution (Dir.) with the Explore-then-Act (E-t-A) paradigm. The results yield several key observations.

\textbf{Exploration-aware training improves performance in both execution paradigms.}
GRPO (Interleaved) consistently outperforms the GRPO (Task-Only) baseline under both the Direct Execution and E-t-A settings. Notably, even though GRPO (Explore-Only) is not explicitly optimized for task execution, it still achieves performance improvements over the base model. This suggests that incorporating exploration-centric rewards during training not only develops exploratory skills, but also enhances the agent’s underlying task-solving capability. In particular, the gains observed under Direct Execution indicate that exploration-aware training encourages a more robust understanding of the environment, which translates into better decision-making even when no separate exploration phase is provided.

\textbf{Exploration-aware training is crucial for realizing the benefits of the E-t-A paradigm.}
Models trained with exploration-specific rewards exhibit consistent improvements when provided with an initial exploration phase. GRPO (Interleaved) and GRPO(Explore-only) achieves positive E-t-A gains across all three environments and both backbone models. This suggests that exploration-focused training enables agents to more effectively convert an exploration budget into actionable information. In contrast, GRPO (Task-Only) exhibits minimal or negative gains in most cases, indicating that conventional task-oriented training does not reliably equip agents with the ability to exploit a separate exploration stage. Together, these results indicate that the E-t-A paradigm is most effective when paired with objectives that explicitly train agents to explore and utilize the information they collect.

\vspace{-0.3em}

\subsection{Analysis}
\label{subsec:analysis}

\begin{figure}[t]
  \centering
\begin{minipage}[t]{0.45\linewidth}
    \vspace{0pt}
    \centering
    \scriptsize
    \setlength{\tabcolsep}{2pt}
    \renewcommand{\arraystretch}{0.85}
    \includegraphics[width=\linewidth]{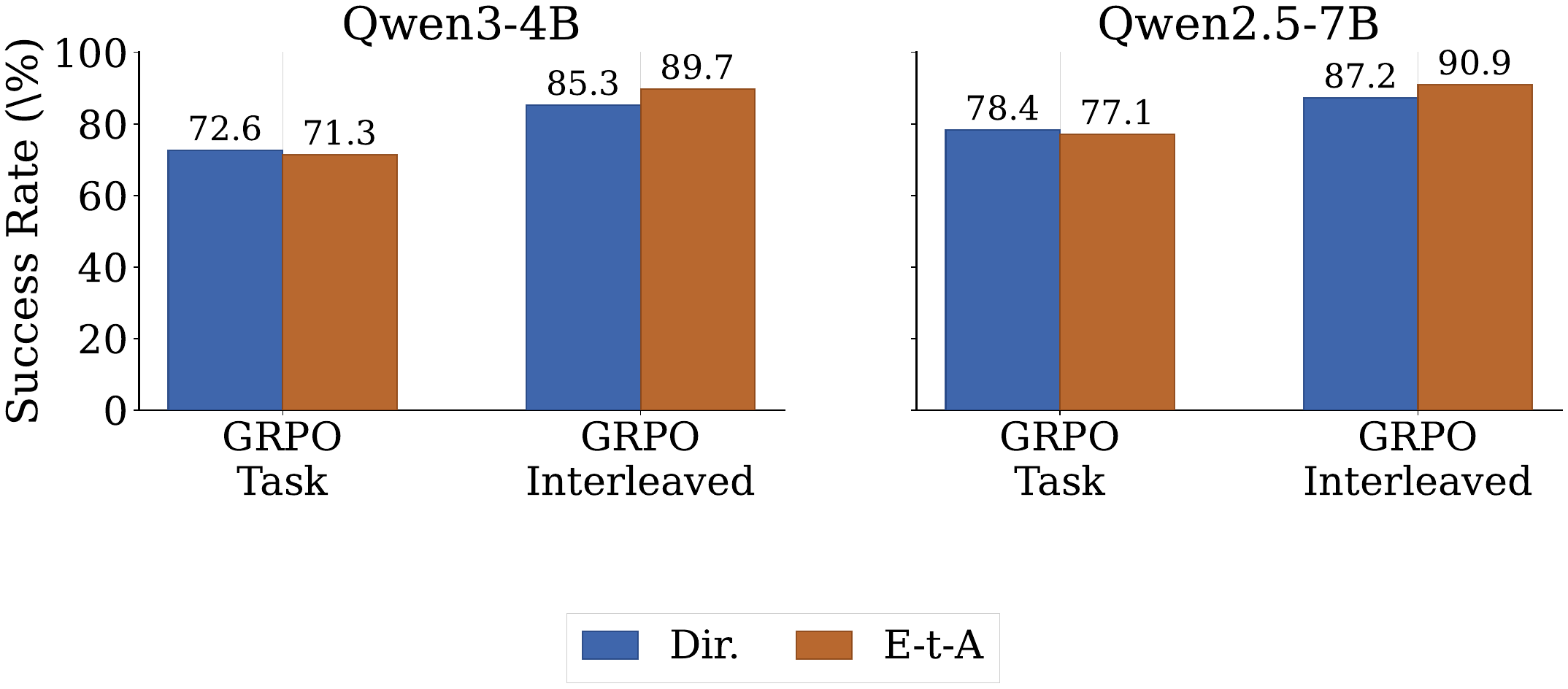}
    \captionof{figure}{
      \textbf{Task Performance on ALFWorld Task Variants}. Exploration-aware training improves adaptation to ALFWorld variant environments, with E-t-A further enhancing the adaptability of the model.
    }
    \label{fig:alfworldvariant}
   
    \renewcommand{\arraystretch}{1.0}
\end{minipage}
  \hfill
  \begin{minipage}[t]{0.52\linewidth}
    \vspace{0pt}
    \centering
    \includegraphics[width=\linewidth]{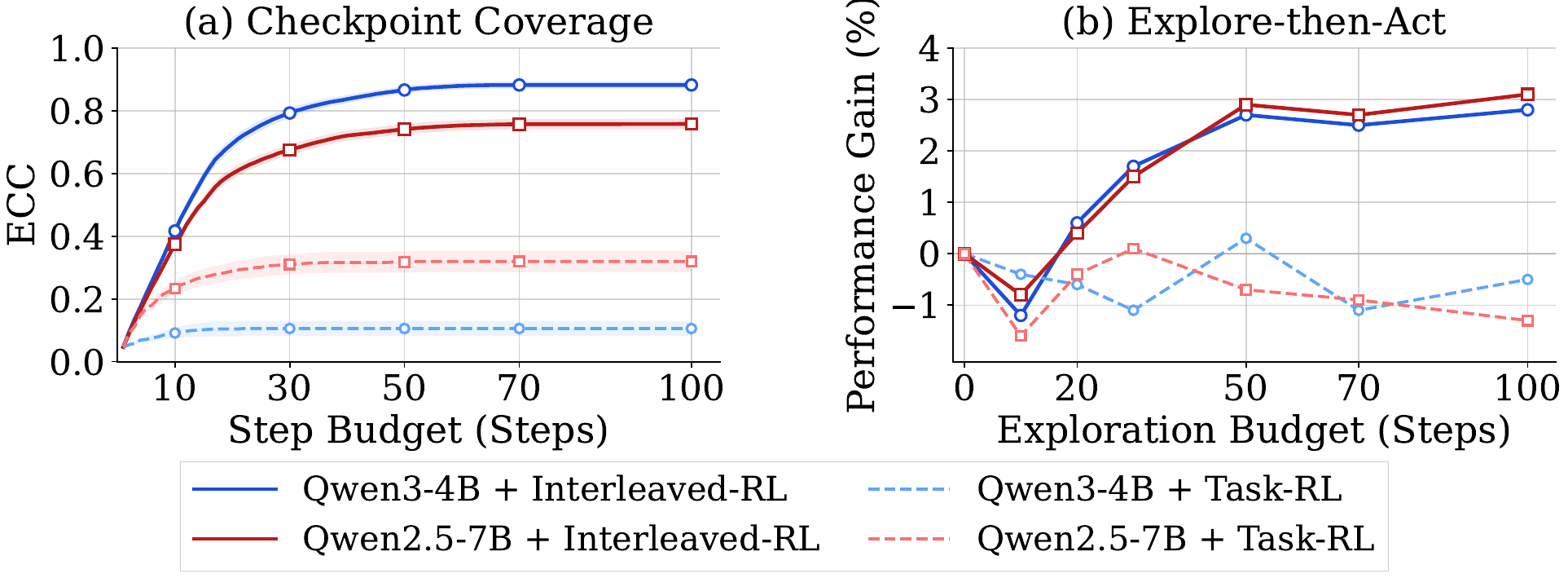}
\captionof{figure}{
  \textbf{Exploration efficiency and downstream task performance on ALFWorld.}
  (a) Environment Checkpoint Coverage (ECC) discovered within a $k$-step budget.
  (b) Explore-then-Act performance gains (\%) over a Qwen3-4B executor baseline
  (30.9\%) when using different models as explorers under a $k$-step exploration budget.}
    \label{fig:ecc-budget}
  \end{minipage}
  \vspace{-0.5em}
\end{figure}

\begin{wraptable}{r}{0.48\textwidth}
  \centering
  \vspace{-1.2em}
  \caption{Direct execution behavior diagnostics.}
  \label{tab:direct_diagnostics}
  \resizebox{0.48\textwidth}{!}{%
    \begin{tabular}{lcc}
      \toprule
      Metric & Task-Only & Exploration-Aware \\
      \midrule
      Repeated Action Rate $\downarrow$ & 63.4\% & \textbf{24.9\%} \\
      Loop Rate $\downarrow$            & 16.0\% & \textbf{7.7\%}  \\
      Info-Seeking Rate $\uparrow$      & 1.0\%  & \textbf{7.5\%}  \\
      Error Recovery Rate $\uparrow$    & 0.0\%  & \textbf{20.1\%} \\
      \bottomrule
    \end{tabular}%
  }
  \vspace{-1em}
\end{wraptable}

\paragraph{The Intrinsic Benefit of Exploration-Aware Training.}

To elucidate why exploration-augmented training enhances direct execution even in the absence of an explicit test-time exploration phase, we analyze the behavioral diagnostics for fail cases presented in Table~\ref{tab:direct_diagnostics}. The data reveal that the GRPO (Task-Only) baseline is prone to a degenerate failure mode wherein it frequently repeats an invalid action, resulting in high rates of repetition and looping with a negligible capacity for recovery. In contrast, agents trained with the GRPO (Interleaved) objective demonstrate a substantial reduction in such repetitive behaviors, concurrently exhibiting an increase in information-seeking and error-recovery actions. These findings indicate that exploration-aware training conditions the model to verify environmental states, dynamically adapt to negative feedback, and pursue alternative strategies, rather than relying on the memorization of rigid action trajectories.  We provide additional case studies in Appendix~\ref{sec:appendix:casestudy}.

\paragraph{Robustness to Perturbations.}
We further investigate the role of autonomous exploration in enhancing agent robustness against environmental shifts. To this end, we introduce perturbed variants of the ALFWorld environment, with modifications to object locations, interaction preconditions, and distractor objects; details are provided in Appendix~\ref{sec:appendix:alfworldvariant}. As illustrated in Figure~\ref{fig:alfworldvariant}, the performance of the task-only model degrades substantially under these perturbations. In contrast, exploration-aware training significantly mitigates this performance degradation. The GRPO (Interleaved) model, when coupled with the E-t-A paradigm, not only achieves the highest success rate on these perturbed tasks but also exhibits the smallest performance decline. This finding indicates that exploration provides an effective mechanism for online adaptation to environmental changes.

\paragraph{Exploration Efficiency and Its Impact on Task Performance.}
To disentangle exploration quality from task-execution proficiency, we analyze how different training objectives influence exploration efficiency. In this experiment, we employ agents trained with GRPO (Task-Only) and GRPO (Interleaved) to serve as dedicated explorers. The exploration knowledge, collected within a budget of $k$ steps, is subsequently provided to a fixed executor agent (the base Qwen3-4B model), which then attempts the task. As shown in Figure~\ref{fig:ecc-budget}(a), the agent trained via GRPO (Interleaved) is a more efficient explorer, achieving higher ECC scores at all budget levels than its task-only counterpart. Figure~\ref{fig:ecc-budget}(b) further demonstrates that this superior exploration quality directly translates into improved downstream task performance. Notably, at a very low budget ($k=10$), the information from both explorers results in a performance decline, confirming that insufficient exploration can introduce counterproductive noise that hinders rather than aids the executor.

\vspace{-0.3em}
\section{Conclusion}
We identify autonomous environment exploration as a missing but essential capability for LLM agents: models optimized primarily for task completion often exhibit premature exploitation. To study this capability systematically, we formalize exploration as an independent and trainable objective, and introduce Exploration Checkpoint Coverage (ECC) as a verifiable metric for quantifying the extent to which agents discover critical states, objects, and affordances within an environment. We further show that exploration can be explicitly instilled through interleaved GRPO with ECC-based rewards, enabling agents for more robust task execution and to first build grounded environment knowledge and then use it for downstream task execution under the Explore-then-Act paradigm. Across diverse interactive environments, our experiments show that naive exploration can hurt performance, whereas exploration-aware training consistently improves both direct task execution and Explore-then-Act performance, highlighting autonomous exploration as a practical path toward more adaptive and generalizable agents.

\bibliography{reference}

@inproceedings{react_paradigm,
  author       = {Shunyu Yao and
                  Jeffrey Zhao and
                  Dian Yu and
                  Nan Du and
                  Izhak Shafran and
                  Karthik R. Narasimhan and
                  Yuan Cao},
  title        = {ReAct: Synergizing Reasoning and Acting in Language Models},
  booktitle    = {The Eleventh International Conference on Learning Representations,
                  {ICLR} 2023, Kigali, Rwanda, May 1-5, 2023},
  publisher    = {OpenReview.net},
  year         = {2023},
  url          = {https://openreview.net/forum?id=WE\_vluYUL-X},
  bibsource    = {dblp computer science bibliography, https://dblp.org}
}

@inproceedings{reflexion_paradigm,
 author = {Shinn, Noah and Cassano, Federico and Gopinath, Ashwin and Narasimhan, Karthik and Yao, Shunyu},
 booktitle = {Advances in Neural Information Processing Systems},
 editor = {A. Oh and T. Naumann and A. Globerson and K. Saenko and M. Hardt and S. Levine},
 pages = {8634--8652},
 publisher = {Curran Associates, Inc.},
 title = {Reflexion: language agents with verbal reinforcement learning},
 url = {https://proceedings.neurips.cc/paper_files/paper/2023/file/1b44b878bb782e6954cd888628510e90-Paper-Conference.pdf},
 volume = {36},
 year = {2023}
}

@misc{agenttuning,
      title={AgentTuning: Enabling Generalized Agent Abilities for LLMs},
      author={Aohan Zeng and Mingdao Liu and Rui Lu and Bowen Wang and Xiao Liu and Yuxiao Dong and Jie Tang},
      year={2023},
      eprint={2310.12823},
      archivePrefix={arXiv},
      primaryClass={cs.CL}
}

@inproceedings{
toolllm,
title={Tool{LLM}: Facilitating Large Language Models to Master 16000+ Real-world {API}s},
author={Yujia Qin and Shihao Liang and Yining Ye and Kunlun Zhu and Lan Yan and Yaxi Lu and Yankai Lin and Xin Cong and Xiangru Tang and Bill Qian and Sihan Zhao and Lauren Hong and Runchu Tian and Ruobing Xie and Jie Zhou and Mark Gerstein and dahai li and Zhiyuan Liu and Maosong Sun},
booktitle={The Twelfth International Conference on Learning Representations},
year={2024},
url={https://openreview.net/forum?id=dHng2O0Jjr}
}

@inproceedings{
gorilla,
title={Gorilla: Large Language Model Connected with Massive {API}s},
author={Shishir G Patil and Tianjun Zhang and Xin Wang and Joseph E. Gonzalez},
booktitle={The Thirty-eighth Annual Conference on Neural Information Processing Systems},
year={2024},
url={https://openreview.net/forum?id=tBRNC6YemY}
}

@misc{agentgym-rl,
      title={AgentGym-RL: Training LLM Agents for Long-Horizon Decision Making through Multi-Turn Reinforcement Learning}, 
      author={Zhiheng Xi and Jixuan Huang and Chenyang Liao and Baodai Huang and Honglin Guo and Jiaqi Liu and Rui Zheng and Junjie Ye and Jiazheng Zhang and Wenxiang Chen and Wei He and Yiwen Ding and Guanyu Li and Zehui Chen and Zhengyin Du and Xuesong Yao and Yufei Xu and Jiecao Chen and Tao Gui and Zuxuan Wu and Qi Zhang and Xuanjing Huang and Yu-Gang Jiang},
      year={2025},
      eprint={2509.08755},
      archivePrefix={arXiv},
      primaryClass={cs.LG},
      url={https://arxiv.org/abs/2509.08755}, 
}

@article{gigpo,
  title={Group-in-Group Policy Optimization for LLM Agent Training},
  author={Feng, Lang and Xue, Zhenghai and Liu, Tingcong and An, Bo},
  journal={arXiv preprint arXiv:2505.10978},
  year={2025}
}

@inproceedings{
realworld,
title={A Real-World WebAgent with Planning, Long Context Understanding, and Program Synthesis},
author={Izzeddin Gur and Hiroki Furuta and Austin V Huang and Mustafa Safdari and Yutaka Matsuo and Douglas Eck and Aleksandra Faust},
booktitle={The Twelfth International Conference on Learning Representations},
year={2024},
url={https://openreview.net/forum?id=9JQtrumvg8}
}

@inproceedings{
swebench,
title={{SWE}-bench: Can Language Models Resolve Real-world Github Issues?},
author={Carlos E Jimenez and John Yang and Alexander Wettig and Shunyu Yao and Kexin Pei and Ofir Press and Karthik R Narasimhan},
booktitle={The Twelfth International Conference on Learning Representations},
year={2024},
url={https://openreview.net/forum?id=VTF8yNQM66}
}

@misc{catastrophicforgetting,
      title={An Empirical Study of Catastrophic Forgetting in Large Language Models During Continual Fine-tuning}, 
      author={Yun Luo and Zhen Yang and Fandong Meng and Yafu Li and Jie Zhou and Yue Zhang},
      year={2025},
      eprint={2308.08747},
      archivePrefix={arXiv},
      primaryClass={cs.CL},
      url={https://arxiv.org/abs/2308.08747}, 
}

@misc{agentrl,
      title={AgentRL: Scaling Agentic Reinforcement Learning with a Multi-Turn, Multi-Task Framework}, 
      author={Hanchen Zhang and Xiao Liu and Bowen Lv and Xueqiao Sun and Bohao Jing and Iat Long Iong and Zhenyu Hou and Zehan Qi and Hanyu Lai and Yifan Xu and Rui Lu and Hongning Wang and Jie Tang and Yuxiao Dong},
      year={2025},
      eprint={2510.04206},
      archivePrefix={arXiv},
      primaryClass={cs.AI},
      url={https://arxiv.org/abs/2510.04206}, 
}

@inproceedings{
hgpo,
title={Hierarchy-of-Groups Policy Optimization for Long-Horizon Agentic Tasks},
author={Shuo He and Lang Feng and Qi Wei and Xin Cheng and Lei Feng and Bo An},
booktitle={The Fourteenth International Conference on Learning Representations},
year={2026},
url={https://openreview.net/forum?id=T8Dev99qnz}
}

@misc{osatlas,
      title={OS-ATLAS: A Foundation Action Model for Generalist GUI Agents}, 
      author={Zhiyong Wu and Zhenyu Wu and Fangzhi Xu and Yian Wang and Qiushi Sun and Chengyou Jia and Kanzhi Cheng and Zichen Ding and Liheng Chen and Paul Pu Liang and Yu Qiao},
      year={2024},
      eprint={2410.23218},
      archivePrefix={arXiv},
      primaryClass={cs.CL},
      url={https://arxiv.org/abs/2410.23218}, 
}

@misc{codeagent,
      title={CodeAgent: Enhancing Code Generation with Tool-Integrated Agent Systems for Real-World Repo-level Coding Challenges}, 
      author={Kechi Zhang and Jia Li and Ge Li and Xianjie Shi and Zhi Jin},
      year={2024},
      eprint={2401.07339},
      archivePrefix={arXiv},
      primaryClass={cs.SE},
      url={https://arxiv.org/abs/2401.07339}, 
}

@article{
voyager,
title={Voyager: An Open-Ended Embodied Agent with Large Language Models},
author={Guanzhi Wang and Yuqi Xie and Yunfan Jiang and Ajay Mandlekar and Chaowei Xiao and Yuke Zhu and Linxi Fan and Anima Anandkumar},
journal={Transactions on Machine Learning Research},
issn={2835-8856},
year={2024},
url={https://openreview.net/forum?id=ehfRiF0R3a},
note={}
}

@article{jiang2025verltool,
  title={VerlTool: Towards Holistic Agentic Reinforcement Learning with Tool Use},
  author={Jiang, Dongfu and Lu, Yi and Li, Zhuofeng and Lyu, Zhiheng and Nie, Ping and Wang, Haozhe and Su, Alex and Chen, Hui and Zou, Kai and Du, Chao and others},
  journal={arXiv preprint arXiv:2509.01055},
  year={2025}
}

@misc{fundamentals,
      title={Fundamentals of Building Autonomous LLM Agents}, 
      author={Victor de Lamo Castrillo and Habtom Kahsay Gidey and Alexander Lenz and Alois Knoll},
      year={2025},
      eprint={2510.09244},
      archivePrefix={arXiv},
      primaryClass={cs.AI},
      url={https://arxiv.org/abs/2510.09244}, 
}

@misc{Agent-R,
      title={Agent-R: Training Language Model Agents to Reflect via Iterative Self-Training}, 
      author={Siyu Yuan and Zehui Chen and Zhiheng Xi and Junjie Ye and Zhengyin Du and Jiecao Chen},
      year={2025},
      eprint={2501.11425},
      archivePrefix={arXiv},
      primaryClass={cs.AI},
      url={https://arxiv.org/abs/2501.11425}, 
}

@misc{mcpatlas,
      title={MCP-Atlas: A Large-Scale Benchmark for Tool-Use Competency with Real MCP Servers}, 
      author={Chaithanya Bandi and Ben Hertzberg and Geobio Boo and Tejas Polakam and Jeff Da and Sami Hassaan and Manasi Sharma and Andrew Park and Ernesto Hernandez and Dan Rambado and Ivan Salazar and Rafael Cruz and Chetan Rane and Ben Levin and Brad Kenstler and Bing Liu},
      year={2026},
      eprint={2602.00933},
      archivePrefix={arXiv},
      primaryClass={cs.SE},
      url={https://arxiv.org/abs/2602.00933}, 
}

@article{osworld,
 author = {Tianbao Xie and
Danyang Zhang and
Jixuan Chen and
Xiaochuan Li and
Siheng Zhao and
Ruisheng Cao and
Toh Jing Hua and
Zhoujun Cheng and
Dongchan Shin and
Fangyu Lei and
Yitao Liu and
Yiheng Xu and
Shuyan Zhou and
Silvio Savarese and
Caiming Xiong and
Victor Zhong and
Tao Yu},
 bibsource = {dblp computer science bibliography, https://dblp.org},
 doi = {10.48550/ARXIV.2404.07972},
 journal = {Advances in Neural Information Processing Systems},
 title = {OSWorld: Benchmarking Multimodal Agents for Open-Ended Tasks in Real
Computer Environments},
 url = {https://doi.org/10.48550/arXiv.2404.07972},
 year = {2024}
}

@article{WebArena,
  title={WebArena: A Realistic Web Environment for Building Autonomous Agents},
  author={Shuyan Zhou and Frank F. Xu and Hao Zhu and Xuhui Zhou and Robert Lo and Abishek Sridhar and Xianyi Cheng and Yonatan Bisk and Daniel Fried and Uri Alon and Graham Neubig},
  journal={ArXiv},
  year={2023},
  volume={abs/2307.13854},
  url={https://api.semanticscholar.org/CorpusID:260164780}
}

@inproceedings{
agentbench,
title={AgentBench: Evaluating {LLM}s as Agents},
author={Xiao Liu and Hao Yu and Hanchen Zhang and Yifan Xu and Xuanyu Lei and Hanyu Lai and Yu Gu and Hangliang Ding and Kaiwen Men and Kejuan Yang and Shudan Zhang and Xiang Deng and Aohan Zeng and Zhengxiao Du and Chenhui Zhang and Sheng Shen and Tianjun Zhang and Yu Su and Huan Sun and Minlie Huang and Yuxiao Dong and Jie Tang},
booktitle={The Twelfth International Conference on Learning Representations},
year={2024},
url={https://openreview.net/forum?id=zAdUB0aCTQ}
}

@misc{tau2,
      title={$\tau^2$-Bench: Evaluating Conversational Agents in a Dual-Control Environment}, 
      author={Victor Barres and Honghua Dong and Soham Ray and Xujie Si and Karthik Narasimhan},
      year={2025},
      eprint={2506.07982},
      archivePrefix={arXiv},
      primaryClass={cs.AI},
      url={https://arxiv.org/abs/2506.07982}, 
}

@misc{cues,
  title         = {CuES: A Curiosity-driven and Environment-grounded Synthesis Framework for Agentic RL},
  author        = {Mai, Shinji and Zhai, Yunpeng and Chen, Ziqian and Chen, Cheng and Zou, Anni and Tao, Shuchang and Liu, Zhaoyang and Ding, Bolin},
  year          = {2025},
  month         = dec,
  eprint        = {2512.01311},
  archivePrefix = {arXiv},
  primaryClass  = {cs.AI},
  doi           = {10.48550/arXiv.2512.01311},
  url           = {https://arxiv.org/abs/2512.01311}
}

@inproceedings{
learnbyinteract,
title={Learn-by-interact: A Data-Centric Framework For Self-Adaptive Agents in Realistic Environments},
author={Hongjin SU and Ruoxi Sun and Jinsung Yoon and Pengcheng Yin and Tao Yu and Sercan O Arik},
booktitle={The Thirteenth International Conference on Learning Representations},
year={2025},
url={https://openreview.net/forum?id=3UKOzGWCVY}
}

@inproceedings{explorer,
    title = "Explorer: Scaling Exploration-driven Web Trajectory Synthesis for Multimodal Web Agents",
    author = "Pahuja, Vardaan  and
      Lu, Yadong  and
      Rosset, Corby  and
      Gou, Boyu  and
      Mitra, Arindam  and
      Whitehead, Spencer  and
      Su, Yu  and
      Awadallah, Ahmed Hassan",
    editor = "Che, Wanxiang  and
      Nabende, Joyce  and
      Shutova, Ekaterina  and
      Pilehvar, Mohammad Taher",
    booktitle = "Findings of the Association for Computational Linguistics: ACL 2025",
    month = jul,
    year = "2025",
    address = "Vienna, Austria",
    publisher = "Association for Computational Linguistics",
    url = "https://aclanthology.org/2025.findings-acl.326/",
    doi = "10.18653/v1/2025.findings-acl.326",
    pages = "6300--6323",
    ISBN = "979-8-89176-256-5"
}

@inproceedings{
walle,
title={{WALL}-E: World Alignment by NeuroSymbolic Learning improves World Model-based {LLM} Agents},
author={Siyu Zhou and Tianyi Zhou and Yijun Yang and Guodong Long and Deheng Ye and Jing Jiang and Chengqi Zhang},
booktitle={The Thirty-ninth Annual Conference on Neural Information Processing Systems},
year={2026},
url={https://openreview.net/forum?id=DorAT49sxj}
}

@misc{tttadapt,
      title={Test-Time Adaptation for LLM Agents via Environment Interaction}, 
      author={Arthur Chen and Zuxin Liu and Jianguo Zhang and Akshara Prabhakar and Zhiwei Liu and Shelby Heinecke and Silvio Savarese and Victor Zhong and Caiming Xiong},
      year={2026},
      eprint={2511.04847},
      archivePrefix={arXiv},
      primaryClass={cs.LG},
      url={https://arxiv.org/abs/2511.04847}, 
}

@misc{wese,
      title={WESE: Weak Exploration to Strong Exploitation for LLM Agents}, 
      author={Xu Huang and Weiwen Liu and Xiaolong Chen and Xingmei Wang and Defu Lian and Yasheng Wang and Ruiming Tang and Enhong Chen},
      year={2024},
      eprint={2404.07456},
      archivePrefix={arXiv},
      primaryClass={cs.AI},
      url={https://arxiv.org/abs/2404.07456}, 
}

@inproceedings{
automanual,
title={AutoManual: Generating Instruction Manuals by {LLM} Agents via Interactive Environmental Learning},
author={Minghao Chen and Yihang Li and Yanting Yang and Shiyu Yu and Binbin Lin and Xiaofei He},
booktitle={The Thirty-eighth Annual Conference on Neural Information Processing Systems},
year={2024},
url={https://openreview.net/forum?id=Pwl9n4zlf5}
}

@inproceedings{
vitabench,
title={VitaBench: Benchmarking {LLM} Agents with Versatile Interactive Tasks in Real-world Applications},
author={Wei He and Yueqing Sun and Hongyan Hao and Xueyuan Hao and Zhikang Xia and Qi GU and Hui Su and Xunliang Cai},
booktitle={The Fourteenth International Conference on Learning Representations},
year={2026},
url={https://openreview.net/forum?id=rtcX9qOBaz}
}

@misc{envscaler,
      title={EnvScaler: Scaling Tool-Interactive Environments for LLM Agent via Programmatic Synthesis}, 
      author={Xiaoshuai Song and Haofei Chang and Guanting Dong and Yutao Zhu and Ji-Rong Wen and Zhicheng Dou},
      year={2026},
      eprint={2601.05808},
      archivePrefix={arXiv},
      primaryClass={cs.CL},
      url={https://arxiv.org/abs/2601.05808}, 
}

@misc{browsecomp,
      title={BrowseComp: A Simple Yet Challenging Benchmark for Browsing Agents}, 
      author={Jason Wei and Zhiqing Sun and Spencer Papay and Scott McKinney and Jeffrey Han and Isa Fulford and Hyung Won Chung and Alex Tachard Passos and William Fedus and Amelia Glaese},
      year={2025},
      eprint={2504.12516},
      archivePrefix={arXiv},
      primaryClass={cs.CL},
      url={https://arxiv.org/abs/2504.12516}, 
}

@inproceedings{
alfworld,
title={{\{}ALFW{\}}orld: Aligning Text and Embodied Environments for Interactive Learning},
author={Mohit Shridhar and Xingdi Yuan and Marc-Alexandre Cote and Yonatan Bisk and Adam Trischler and Matthew Hausknecht},
booktitle={International Conference on Learning Representations},
year={2021},
url={https://openreview.net/forum?id=0IOX0YcCdTn}
}

@inproceedings{scienceworld,
    title = "{S}cience{W}orld: Is your Agent Smarter than a 5th Grader?",
    author = "Wang, Ruoyao  and
      Jansen, Peter  and
      C{\^o}t{\'e}, Marc-Alexandre  and
      Ammanabrolu, Prithviraj",
    editor = "Goldberg, Yoav  and
      Kozareva, Zornitsa  and
      Zhang, Yue",
    booktitle = "Proceedings of the 2022 Conference on Empirical Methods in Natural Language Processing",
    month = dec,
    year = "2022",
    address = "Abu Dhabi, United Arab Emirates",
    publisher = "Association for Computational Linguistics",
    url = "https://aclanthology.org/2022.emnlp-main.775/",
    doi = "10.18653/v1/2022.emnlp-main.775",
    pages = "11279--11298",
}

@misc{agentgym,
      title={AgentGym: Evolving Large Language Model-based Agents across Diverse Environments}, 
      author={Zhiheng Xi and Yiwen Ding and Wenxiang Chen and Boyang Hong and Honglin Guo and Junzhe Wang and Dingwen Yang and Chenyang Liao and Xin Guo and Wei He and Songyang Gao and Lu Chen and Rui Zheng and Yicheng Zou and Tao Gui and Qi Zhang and Xipeng Qiu and Xuanjing Huang and Zuxuan Wu and Yu-Gang Jiang},
      year={2024},
      eprint={2406.04151},
      archivePrefix={arXiv},
      primaryClass={cs.AI}
}

@article{Qwen3,
  author       = {An Yang and
                  Anfeng Li and
                  Baosong Yang and
                  Beichen Zhang and
                  Binyuan Hui and
                  Bo Zheng and
                  Bowen Yu and
                  Chang Gao and
                  Chengen Huang and
                  Chenxu Lv and
                  Chujie Zheng and
                  Dayiheng Liu and
                  Fan Zhou and
                  Fei Huang and
                  Feng Hu and
                  Hao Ge and
                  Haoran Wei and
                  Huan Lin and
                  Jialong Tang and
                  Jian Yang and
                  Jianhong Tu and
                  Jianwei Zhang and
                  Jian Yang and
                  Jiaxi Yang and
                  Jingren Zhou and
                  Junyang Lin and
                  Kai Dang and
                  Keqin Bao and
                  Kexin Yang and
                  Le Yu and
                  Lianghao Deng and
                  Mei Li and
                  Mingfeng Xue and
                  Mingze Li and
                  Pei Zhang and
                  Peng Wang and
                  Qin Zhu and
                  Rui Men and
                  Ruize Gao and
                  Shixuan Liu and
                  Shuang Luo and
                  Tianhao Li and
                  Tianyi Tang and
                  Wenbiao Yin and
                  Xingzhang Ren and
                  Xinyu Wang and
                  Xinyu Zhang and
                  Xuancheng Ren and
                  Yang Fan and
                  Yang Su and
                  Yichang Zhang and
                  Yinger Zhang and
                  Yu Wan and
                  Yuqiong Liu and
                  Zekun Wang and
                  Zeyu Cui and
                  Zhenru Zhang and
                  Zhipeng Zhou and
                  Zihan Qiu},
  title        = {Qwen3 Technical Report},
  journal      = {CoRR},
  volume       = {abs/2505.09388},
  year         = {2025}
}

@misc{claudeopus,
  author       = {{Anthropic}},
  title        = {Introducing Claude Opus 4.5},
  year         = {2025},
  month        = nov,
  day          = {24},
  howpublished = {\url{https://www.anthropic.com/news/claude-opus-4-5}},
  note         = {Accessed: 2026-04-29}
}

@article{GPT4,
  author       = {OpenAI},
  title        = {{GPT-4} Technical Report},
  journal      = {CoRR},
  volume       = {abs/2303.08774},
  year         = {2023}
}

@article{llama1,
  author       = {Hugo Touvron and
                  Thibaut Lavril and
                  Gautier Izacard and
                  Xavier Martinet and
                  Marie{-}Anne Lachaux and
                  Timoth{\'{e}}e Lacroix and
                  Baptiste Rozi{\`{e}}re and
                  Naman Goyal and
                  Eric Hambro and
                  Faisal Azhar and
                  Aur{\'{e}}lien Rodriguez and
                  Armand Joulin and
                  Edouard Grave and
                  Guillaume Lample},
  title        = {LLaMA: Open and Efficient Foundation Language Models},
  journal      = {CoRR},
  volume       = {abs/2302.13971},
  year         = {2023}
}

@article{Qwen2.5,
  author       = {An Yang and
                  Baosong Yang and
                  Beichen Zhang and
                  Binyuan Hui and
                  Bo Zheng and
                  Bowen Yu and
                  Chengyuan Li and
                  Dayiheng Liu and
                  Fei Huang and
                  Haoran Wei and
                  Huan Lin and
                  Jian Yang and
                  Jianhong Tu and
                  Jianwei Zhang and
                  Jianxin Yang and
                  Jiaxi Yang and
                  Jingren Zhou and
                  Junyang Lin and
                  Kai Dang and
                  Keming Lu and
                  Keqin Bao and
                  Kexin Yang and
                  Le Yu and
                  Mei Li and
                  Mingfeng Xue and
                  Pei Zhang and
                  Qin Zhu and
                  Rui Men and
                  Runji Lin and
                  Tianhao Li and
                  Tingyu Xia and
                  Xingzhang Ren and
                  Xuancheng Ren and
                  Yang Fan and
                  Yang Su and
                  Yichang Zhang and
                  Yu Wan and
                  Yuqiong Liu and
                  Zeyu Cui and
                  Zhenru Zhang and
                  Zihan Qiu},
  title        = {Qwen2.5 Technical Report},
  journal      = {CoRR},
  volume       = {abs/2412.15115},
  year         = {2024}
}

@article{ragen,
  author       = {Zihan Wang and
                  Kangrui Wang and
                  Qineng Wang and
                  Pingyue Zhang and
                  Linjie Li and
                  Zhengyuan Yang and
                  Xing Jin and
                  Kefan Yu and
                  Minh Nhat Nguyen and
                  Licheng Liu and
                  Eli Gottlieb and
                  Yiping Lu and
                  Kyunghyun Cho and
                  Jiajun Wu and
                  Li Fei{-}Fei and
                  Lijuan Wang and
                  Yejin Choi and
                  Manling Li},
  title        = {{RAGEN:} Understanding Self-Evolution in {LLM} Agents via Multi-Turn
                  Reinforcement Learning},
  journal      = {CoRR},
  volume       = {abs/2504.20073},
  year         = {2025}
}
\bibliographystyle{unsrt}
\newpage
\appendix

\section{Limitations and Future work.}
\label{sec:appendix:limitation}

Our work takes an initial step toward incentivizing autonomous exploration abilities in LLM-based agents. Looking ahead, we consider following potential limitations and future work.
First, this work studies exploration primarily as an initial phase before task execution, providing a clean and controllable setting for isolating, measuring, and training exploration ability. However, real-world environments are often too large and complex to be fully explored upfront. Extending our framework to dynamic, task-conditioned exploration is therefore an important direction for future work.
Second, our experiments focus on text-based interactive environments, where language provides clear affordances and enables verifiable coverage metrics for studying exploration. Extending exploration to more open-ended multimodal environments is another promising direction.
Overall, we view this work as a foundation for a broader research agenda on exploration-capable agents. Future progress on dynamic and multimodal exploration may further enable agents to acquire grounded environment knowledge online, adapt to evolving conditions, and operate robustly in realistic deployment settings.

\section{Boarder Impact}
\label{sec:boardimpact}
This work formalizes autonomous exploration as a measurable capability for LLM agents and introduces training strategies to improve it. By enabling agents to acquire grounded environment knowledge online, our methods may benefit applications such as virtual assistants, web automation, educational tools, and embodied AI systems, especially in unfamiliar or changing environments.
While this work is methodological and does not directly deploy real-world agents, stronger exploration ability may indirectly increase agent autonomy. Agents that better discover tools, rules, and affordances could also interact with environments in unexpected ways. Therefore, practical deployment should include appropriate safeguards, such as permission control, monitoring, constrained environments, and human oversight for high-stakes actions.
Overall, this work provides evaluation and training tools for building more adaptive and generalizable LLM agents, while highlighting the need for responsible use as autonomous exploration capabilities improve.

\section{Addtional Experimental Details}
\label{sec:appendix:addexpdetails}

\paragraph{Group Relative Policy Optimization (GRPO).}
For GRPO training, we follow the formulation described in Section~\ref{subsec:method:training}. In the interleaved training setting, each batch contains both task-execution rollouts (rewarded by binary task success) and exploration rollouts (rewarded by ECC). By default, we use a 5:1 ratio of task-execution to exploration rollouts per training batch. Table~\ref{tab:grpo_hyperparams} lists the GRPO-specific hyperparameters.

\begin{table}[h]
    \centering
    \caption{Hyperparameters for GRPO training.}
    \label{tab:grpo_hyperparams}
    \resizebox{0.6\textwidth}{!}{
    \begin{tabular}{lcc}
        \toprule
        \textbf{Hyperparameter} & \textbf{Qwen2.5-7B} & \textbf{Qwen3-4B} \\
        \midrule
        Learning Rate & 1e-6 & 1e-6 \\
        LR Scheduler & constant & constant \\
        Group Size ($G$) & 8 & 8 \\
        Max Training Step & 300 & 300 \\
        Max Rollout Length & 32k & 32k \\
        Batch Size & 16 & 16 \\
        Rollouts Per Step & 128 & 128 \\
        \bottomrule
    \end{tabular}
    }
\end{table}

\paragraph{Training Resources.}
All experiments are conducted on a single node equipped with $8 \times$ NVIDIA H800  GPUs. GRPO training requires approximately 192 GPU-hours due to the online rollout generation process. 

\section{Sensitivity to the Task-Exploration Ratio}
\label{sec:appendix:sensitive}

\paragraph{Setup.}
We examine how the balance between task-execution and exploration rollouts affects GRPO training. All runs use Qwen3-4B on ALFWorld with the same training budget and hyperparameters, while varying only the composition of rollouts in each training batch. We include two endpoint baselines, Task-Only and Explore-Only, and six mixed task:exploration ratios: 1:10, 1:5, 1:3, 3:1, 5:1, and 10:1. Each trained policy is evaluated under both Direct Execution (Dir.) and Explore-then-Act (E-t-A), where the latter provides an exploration phase before task execution. Results are reported in Table~\ref{tab:ratio_sensitivity}.

\begin{table}[h]
    \centering
    \caption{Sensitivity to task:exploration ratio in GRPO training. Evaluated on Qwen3-4B, ALFWorld. Dir.\ = Direct Execution success rate (\%), E-t-A = Explore-then-Act success rate (\%).}
    \label{tab:ratio_sensitivity}
    \resizebox{0.55\textwidth}{!}{
    \begin{tabular}{lcc}
        \toprule
        \textbf{Task-Exploration Ratio} & \textbf{Dir. (\%)} & \textbf{E-t-A (\%)} \\
        \midrule
        Explore-Only & $49.7 \pm 2.4$ & $54.4 \pm 2.2$ \\
        1:10         & $60.9 \pm 2.0$ & $64.3 \pm 1.9$ \\
        1:5          & $79.5 \pm 1.6$ & $73.2 \pm 1.8$ \\
        1:3          & $82.7 \pm 1.4$ & $83.9 \pm 0.9$ \\
        3:1          & $86.1 \pm 1.3$ & $88.1 \pm 1.4$ \\
        5:1          & $\mathbf{90.5 \pm 1.1}$ & $\mathbf{92.7 \pm 1.2}$ \\
        10:1         & $87.9 \pm 1.3$ & $91.0 \pm 2.3$ \\
        Task-Only    & $84.6 \pm 1.2$ & $84.3 \pm 1.7$ \\
        \bottomrule
    \end{tabular}
    }
\end{table}

\paragraph{Analysis.}
Task-Only training preserves strong direct task performance but provides almost no benefit from the additional exploration phase, while Explore-Only training substantially underperforms because it lacks enough task-completion signal. Mixed training improves over both endpoints once task rollouts are sufficiently represented. Performance rises as the ratio shifts from exploration-heavy settings toward task-heavy settings, with 5:1 achieving the best Direct and E-t-A success rates. Increasing the task share further to 10:1 slightly reduces performance, suggesting that too little exploration weakens the transferable environment knowledge used by E-t-A. We therefore use 5:1 as the default ratio in the main GRPO experiments, as it provides the best empirical trade-off between task optimization and exploration-aware behavior.

\section{Construction of Environment Checkpoints}
\label{sec:appendix:ECCcondtruction}

\begin{wrapfigure}{r}{0.48\textwidth}
\vspace{-10pt}
\begin{minipage}{0.48\textwidth}
\begin{algorithm}[H]
\caption{ECC Checkpoint Construction}
\label{alg:ecc_construction}
\begin{algorithmic}[1]
\REQUIRE Environment engine $\mathcal{E}$, instance $I$
\ENSURE Checkpoint set $\mathcal{C}$
\STATE $\mathcal{C} \leftarrow \emptyset$
\STATE $\mathcal{S} \leftarrow$ \textsc{GetReachableStates}($\mathcal{E}$, $I$)
\FOR{each state $s \in \mathcal{S}$}
    \STATE $L \leftarrow$ \textsc{ExtractLocations}($s$)
    \STATE $O \leftarrow$ \textsc{ExtractObjects}($s$)
    \STATE $A \leftarrow$ \textsc{ExtractAffordances}($s$)
    \STATE $\mathcal{C} \leftarrow \mathcal{C} \cup L \cup O \cup A$
\ENDFOR
\STATE $\mathcal{C} \leftarrow$ \textsc{Deduplicate}($\mathcal{C}$)
\STATE $\mathcal{C} \leftarrow$ \textsc{FilterByRelevance}($\mathcal{C}$)
\RETURN $\mathcal{C}$
\end{algorithmic}
\end{algorithm}
\end{minipage}
\vspace{-10pt}
\end{wrapfigure}

As described in Section~\ref{subsec:method:ecc}, Exploration Checkpoint Coverage (ECC) requires a predefined set of checkpoints $\mathcal{C} = \{c_1, c_2, \dots, c_M\}$ for each environment instance. Here we detail the construction procedure, which leverages the environment engine's internal state representation to derive verifiable, ground-truth checkpoints without relying on any model-generated annotations.

\paragraph{General Procedure.}
For each environment instance, we extract checkpoints from three categories: (1)~\textbf{Locations}: distinct navigable rooms or areas the agent can visit; (2)~\textbf{Objects}: key interactable entities present in the environment; and (3)~\textbf{Affordances}: valid actions or state transitions associated with specific objects or locations (e.g., an object that can be opened, a device that can be activated). Algorithm~\ref{alg:ecc_construction} provides pseudocode for this extraction pipeline.

\paragraph{Verification Mechanism.}
A checkpoint $c_i$ is marked as \emph{covered} if the agent's exploration trajectory contains an observation or action that unambiguously demonstrates awareness of $c_i$. Specifically, we perform string matching against the environment's textual observations: a location checkpoint is triggered when the agent receives the corresponding room description, an object checkpoint is triggered when the object appears in an observation following an interaction or examination action, and an affordance checkpoint is triggered when the agent successfully executes the associated valid action. This verification is deterministic and does not require any LLM-based judgment.

\paragraph{Environment-Specific Details.}

\begin{itemize}[leftmargin=*, parsep=0pt, itemsep=3pt]
    \item \textbf{ALFWorld.} Checkpoints are derived from the PDDL game state. Locations correspond to navigable rooms (e.g., kitchen, bedroom, bathroom). Objects include all task-relevant items and receptacles. Affordances cover valid pick-up, put, open, close, and toggle actions.

    \item \textbf{ScienceWorld.} Checkpoints are extracted from the environment's object tree and action space. Locations include all accessible rooms in the simulated house and yard. Objects encompass scientific instruments, materials, and containers. Affordances capture valid experimental operations (e.g., heating, mixing, measuring) and state changes (e.g., substance melting, temperature rising).

    \item \textbf{TextCraft.} Checkpoints are derived from the crafting recipe graph. Locations represent distinct resource-gathering zones. Objects include raw materials and intermediate crafted items. Affordances correspond to valid crafting recipes and resource-gathering commands that the agent can execute.
\end{itemize}

\section{Detailed Construction of ALFWorld Variants}
\label{sec:appendix:alfworldvariant}

To evaluate whether exploration-capable agents can adapt to environment shifts at test time, we construct three perturbed variants of ALFWorld. Each variant modifies a single axis of the environment while preserving the underlying task structure, so that performance degradation can be attributed to the agent's inability to handle the specific perturbation type rather than a fundamentally different task. All variants are derived from the same 274 test instances used in the original ALFWorld evaluation, yielding 274 instances per variant (1,096 total including the original).

\paragraph{Variant 1: Object Relocation.}
We modify the initial placement of task-relevant objects and receptacles. Specifically, for each task instance, we randomly reassign target objects to different receptacles or rooms while ensuring that the task remains solvable (i.e., all necessary objects are still reachable). For example, a task that originally requires finding a mug on the kitchen counter may now have the mug placed inside a bedroom drawer. This variant tests whether an agent has memorized fixed object--location associations from training or can discover the current object layout through exploration.

\paragraph{Variant 2: Interaction Precondition Changes.}
We alter the preconditions required to interact with certain objects or receptacles. For example, a container that is normally open by default may now start in a closed state and require an explicit \texttt{open} action before the agent can access its contents, or a receptacle that previously accepted objects directly may now require the agent to first clear an existing item. These modifications change the valid action sequences without altering the spatial layout, testing whether the agent can identify and adapt to new affordance constraints through exploratory interaction.

\paragraph{Variant 3: Distractor Injection.}
We introduce additional distractor objects into the environment that are visually or semantically similar to the task-relevant targets. For instance, in a task requiring the agent to pick up a specific book, we add several additional books in different locations. This variant increases the ambiguity of the environment and tests whether the agent can distinguish the correct target from distractors, a capability that benefits from thorough exploration and environment mapping prior to task execution.

\paragraph{Summary.}
Table~\ref{tab:alfworld_variant_stats} summarizes the statistics of the original and variant ALFWorld test sets.

\begin{table}[h]
    \centering
    \caption{ALFWorld variant statistics.}
    \label{tab:alfworld_variant_stats}
    \begin{tabular}{lc}
        \toprule
        \textbf{Setting} & \textbf{\# Instances} \\
        \midrule
        Original             & 274 \\
        Variant 1: Object Relocation           & 274 \\
        Variant 2: Interaction Precondition     & 274 \\
        Variant 3: Distractor Injection         & 274 \\
        \midrule
        \textbf{Total}       & 1,096 \\
        \bottomrule
    \end{tabular}
\end{table}

\section{Prompt for Exploration}
\label{sec:appendix:prompt}

\begin{tcblisting}{
    colframe=black,
    title=General Exploration Prompt,
    colback=gray!5,
    fonttitle=\bfseries,
    breakable,
    listing only,
    listing options={
        basicstyle=\ttfamily\small,
        breaklines=true,
        columns=fullflexible
    }
}
You are an general curious agent. Your goal is to fully understand the 
environment through systematic exploration.

Objectives
1. Systematically explore all available actions and observe their effects
2. Map out the information structure of the environment
3. Identify reliable patterns and clues

Exploration Strategy
- Try different actions to understand state transitions
- Note what information is available at each state
- Track which actions are reversible vs irreversible
- Identify key decision points
- Explore different paths and branches

IMPORTANT NOTE
All findings must be grounded in direct interaction with the environment, not in prior knowledge or assumptions.
- Do not assume how an action behaves -- execute it and observe the actual result.
- Do not skip exploration because something "seems obvious" -- the environment may violate common expectations.
- When your expectation conflicts with an observation, always trust the observation and update your understanding accordingly.

Output Format
You can output your findings at any time when you feel exploration is complete.
Wrap your world modeling analysis in <world_modeling> </world_modeling> tags.
Describe freely in natural language:
- How actions change the environment
- Where information is located
- Which clues are reliable
- Key insights about the environment structure
- Any other information you found interesting or new
- ...

Example:
<world_modeling>

</world_modeling>

Environment Description
{environment_description_str}

Response Format
Think through your approach step by step, wrap your thinking process in 
<think> </think>
<think>
Your reasoning and analysis here...
</think>
{response_format_str}

You can output world modeling only when you feel exploration is fully enough (Thus at the end of the exploration).
Wrap your world modeling in <world_modeling> </world_modeling> tags.
\end{tcblisting}

\section{Case Study}
\label{sec:appendix:casestudy}

We present two qualitative case studies from ALFWorld to illustrate the behavioral differences between exploration-aware and task-oriented agents. All examples are drawn from real model rollouts on the same environment instances.

\subsection{Task-Oriented Training VS Exploration-Aware Training}

We compare exploration trajectories produced by GRPO~(Mixed) (exploration-aware) and GRPO~(Task-Only) (task-oriented) on the same ALFWorld bedroom environment (task instance: \texttt{pick\_and\_place\_simple}). Both models receive identical goal-free exploration instructions and a budget of 50 interaction steps.

\vspace{6pt}

\begin{promptbox}{Task-Oriented Model (GRPO Task-Only) \textnormal{--- ECC: 0 / Steps: 1}}

\textbf{Initial Observation:}\\
{\small\texttt{You are in the middle of a room. Looking quickly around you, you see a bed~1, a diningtable~1, a drawer~4, a drawer~3, a drawer~2, a drawer~1, a garbagecan~1, a sidetable~2, and a sidetable~1.}}

\vspace{4pt}
\textbf{Step 1:} \texttt{done} \hfill \textcolor{red!70!black}{\textit{(Immediately terminates exploration)}}

\vspace{6pt}
\textbf{Failure Mode: Premature Termination.} The task-oriented model finds no actionable goal in the exploration setting. Without a task reward signal, it immediately outputs \texttt{done} after a single observation, failing to interact with any object or discover any environment dynamics.

\end{promptbox}

\vspace{4pt}

\begin{promptbox}[colframe=red!40!gray, borderline west={3pt}{0pt}{red!50!black}]{Task-Oriented Model: Generated World Model}
\small
Environment is a household setting with various locations and objects. The state space includes:
\begin{itemize}[leftmargin=1.2em, itemsep=1pt, topsep=2pt]
\item Locations: bed 1, diningtable 1, drawer 1--4, sidetable 1--2, garbagecan 1
\item Objects: various items in receptacles (not explicitly named)
\end{itemize}
Key structural features:
\begin{itemize}[leftmargin=1.2em, itemsep=1pt, topsep=2pt]
\item The environment is navigable between locations.
\item Objects are stored in receptacles and can be picked up, moved, examined, or processed.
\item Some objects have specific actions (heat, cool, clean, slice) that modify their state.
\item The robot can carry only one object at a time.
\end{itemize}
\textit{I need to explore the environment to learn more about the specific objects, their properties, and the rules governing interactions.}

\vspace{4pt}
\textcolor{red!60!black}{\textbf{Assessment:} Generic prior knowledge only. No grounded observations --- objects ``not explicitly named,'' action syntax unknown, no verified constraints. The model \textit{acknowledges} it needs to explore but has already terminated.}
\end{promptbox}

\vspace{10pt}

\begin{promptbox}{Exploration-Aware Model (GRPO Mixed) \textnormal{--- ECC: 87\% / Steps: 49}}

\textbf{Initial Observation:} \textit{(Same environment as above)}

\vspace{4pt}
{\small
\begin{tabular}{@{}r@{\;}l@{\enspace}p{7.2cm}@{}}
\multicolumn{3}{@{}l}{\textit{Phase 1: Object Interaction Discovery}} \\[2pt]
1. & \texttt{look} & Observes room layout with all receptacles. \\
2. & \texttt{go to bed 1} & Arrives. Sees: book~1, laptop~1, pillow~2, pillow~1. \\
3. & \texttt{examine book 1} & ``Nothing happens.'' (learns examine needs holding) \\
4. & \texttt{take book 1 from bed 1} & Picks up book~1 successfully. \\
5. & \texttt{inventory} & Confirms: ``You are carrying: a book 1.'' \\
6. & \texttt{examine book 1} & ``There's nothing special about book 1.'' \\
7. & \texttt{take laptop 1 from bed 1} & Fails $\to$ \textit{discovers one-object limit.} \\
8. & \texttt{move book 1 to bed 1} & Places object back. Learns \texttt{move} syntax. \\[4pt]

\multicolumn{3}{@{}l}{\textit{Phase 2: Receptacle Exploration}} \\[2pt]
9. & \texttt{inventory} & Confirms empty hands. \\
10. & \texttt{take laptop 1 from bed 1} & Tests picking different object type. \\
11. & \texttt{move laptop 1 to bed 1} & Returns it. \\
12. & \texttt{go to diningtable 1} & Sees: cd~2, cellphone~3, cellphone~1, mug~1, \ldots \\
13. & \texttt{take mug 1 from diningtable 1} & Picks up mug. \\
14. & \texttt{go to drawer 1} & ``The drawer 1 is closed.'' \\
15. & \texttt{open drawer 1} & Opens drawer $\to$ discovers container mechanics. \\[4pt]

\multicolumn{3}{@{}l}{\quad\textit{\ldots continues visiting sidetable 1--2, drawer 2--4, testing move \ldots}} \\[4pt]

\multicolumn{3}{@{}l}{\textit{Phase 3: Action Verification}} \\[2pt]
45. & \texttt{go to drawer 1} & Returns to verify drawer state persistence. \\
46. & \texttt{open drawer 1} & Re-confirms open/close mechanics. \\
47. & \texttt{close drawer 1} & Tests close action. \\
48. & \texttt{go to bed 1} & Final sweep of starting location. \\
49. & \texttt{done} & Terminates after exhaustive exploration. \\
\end{tabular}
}

\end{promptbox}

\vspace{4pt}

\begin{promptbox}[colframe=blue!40!gray, borderline west={3pt}{0pt}{blue!50!black}]{Exploration-Aware Model: Generated World Model}
\small
\textbf{Room Structure:}
Single room (bedroom type) containing: bed, diningtable, drawers~(1--4), garbagecan, sidetables~(1--2).

\vspace{3pt}
\textbf{Actions and Their Effects:}
\begin{itemize}[leftmargin=1.2em, itemsep=1pt, topsep=2pt]
\item \texttt{look} --- Shows room description; subsequent uses may return less info.
\item \texttt{inventory} --- Shows held object or ``You are not carrying anything.''
\item \texttt{go to (receptacle)} --- Moves to receptacle; shows contents on arrival.
\item \texttt{open/close (receptacle)} --- For containers (drawers, fridges, cabinets).
\item \texttt{take (obj) from (recep)} --- Requires: at receptacle, not holding another.
\item \texttt{move (obj) to (recep)} --- Requires: holding object, at target receptacle.
\item \texttt{examine (object)} --- Only works on held objects.
\item \texttt{heat/cool/clean (obj) with (recep)} --- State-changing with specific devices.
\end{itemize}

\vspace{3pt}
\textbf{Key Constraints (Verified):}
\begin{enumerate}[leftmargin=1.5em, itemsep=1pt, topsep=2pt]
\item \textbf{Single-item inventory:} Exactly one object at a time; second take fails silently.
\item \textbf{Proximity required:} Must be at receptacle for take/move.
\item \textbf{Container state:} Drawers must be opened before contents visible.
\item \textbf{Examine requires holding:} Cannot examine objects on receptacles.
\item \textbf{Error signal:} Invalid actions return ``Nothing happens'' (no error type).
\end{enumerate}

\vspace{3pt}
\textbf{Discovered Objects:}
bed~1: book~1, laptop~1, pillow~1--2; diningtable~1: cd~2, cellphone~1/3, mug~1, keychain~1; drawer~1: pencil~2; sidetable~1--2: various objects.

\vspace{4pt}
\textcolor{blue!60!black}{\textbf{Assessment:} Fully grounded knowledge with verified action semantics, correct syntax, discovered constraints, and complete object inventory.}
\end{promptbox}

\subsection{Low-ECC vs.\ High-ECC Grounded Task Execution}
\label{subsec:casestudy:task}

We compare task execution on the same ALFWorld kitchen environment (task: \emph{pick\_cool\_then\_place\_in\_recep} --- pick up a mug, cool it, then place it on the coffeemachine). The exploration-aware model has access to its rich exploration summary (high ECC), while the task-oriented model operates with its shallow summary (zero ECC).

\vspace{6pt}

\begin{promptbox}{High-ECC Grounded Execution (Exploration-Aware) \textnormal{--- \textcolor{green!60!black}{\cmark} Success / 7 steps}}

\textbf{Task:} Pick up a mug, cool it with the fridge, then place it on the coffeemachine.

\vspace{4pt}
{\small
\begin{tabular}{@{}r@{\;}l@{\enspace}p{7.2cm}@{}}
1. & \texttt{look} & Observes kitchen: cabinet~1--6, coffeemachine~1, countertop~1--3, fridge~1, \ldots \\
2. & \texttt{go to countertop 1} & Arrives. Sees: glassbottle~2, \textbf{mug~1}, papertowelroll~1. \\
3. & \texttt{take mug 1 from countertop 1} & Picks up target object. \\
4. & \texttt{go to fridge 1} & Navigates directly to cooling device. \\
5. & \texttt{cool mug 1 with fridge 1} & Successfully cools the mug. \\
6. & \texttt{go to coffeemachine 1} & Navigates to target placement. \\
7. & \texttt{move mug 1 to coffeemachine 1} & Places cooled mug. \textbf{Task Complete!} \\
\end{tabular}
}

\vspace{4pt}
\textcolor{green!60!black}{\textbf{Key:}} The agent knows the exact syntax (\texttt{cool X with fridge 1}), the one-object constraint, and efficiently locates the mug on its first receptacle visit.

\end{promptbox}

\vspace{8pt}

\begin{promptbox}{Zero-ECC Grounded Execution (Task-Only) \textnormal{--- \textcolor{red!70!black}{\xmark} Failure / 100 steps}}

\textbf{Task:} \textit{(Same task, same environment as above)}

\vspace{4pt}
{\small
\begin{tabular}{@{}r@{\;}l@{\enspace}p{7.2cm}@{}}
\multicolumn{3}{@{}l}{\textit{Phase 1: Aimless search (steps 1--10)}} \\[2pt]
1. & \texttt{look} & Observes kitchen (same environment). \\
2. & \texttt{go to fridge 1} & Goes to fridge first (no plan). \\
3. & \texttt{open fridge 1} & Opens fridge. Sees: apple, bowl, egg --- \textit{no mug.} \\
4. & \texttt{go to cabinet 1} & Starts searching cabinets. Sees: vase~1. \\
5. & \texttt{go to cabinet 2} & Opens cabinet~2. Sees: glassbottle, saltshaker, vase. \\
6--10. & \texttt{go to cabinet 3--5} & Continues cabinets. Finds plates, saltshaker, soap. \\[4pt]

\multicolumn{3}{@{}l}{\quad\textit{\ldots (steps 11--60: exhaustively searches all cabinets, drawers, sinkbasin;}} \\
\multicolumn{3}{@{}l}{\quad\textit{eventually finds mug but fails to cool due to incorrect sequencing) \ldots}} \\[4pt]

\multicolumn{3}{@{}l}{\textit{Phase 2: Stuck in loop (steps 96--100)}} \\[2pt]
96. & \texttt{cool mug 1 with fridge 1} & \textcolor{red!70!black}{``Nothing happens.''} (not at fridge) \\
97. & \texttt{cool mug 1 with fridge 1} & \textcolor{red!70!black}{``Nothing happens.''} \\
98. & \texttt{cool mug 1 with fridge 1} & \textcolor{red!70!black}{``Nothing happens.''} \\
99. & \texttt{cool mug 1 with fridge 1} & \textcolor{red!70!black}{``Nothing happens.''} \\
100. & \texttt{cool mug 1 with fridge 1} & \textcolor{red!70!black}{``Nothing happens.''} $\to$ \textbf{Budget exhausted.} \\
\end{tabular}
}

\vspace{4pt}
\textcolor{red!70!black}{\textbf{Failure mode:}} The agent (1)~searches inefficiently without knowing where mugs are located, (2)~attempts the correct action but violates the proximity constraint, and (3)~perseverates on the same failed action without adapting.

\end{promptbox}

\vspace{8pt}

\paragraph{Summary.}
These case studies illustrate two complementary findings. First, exploration-aware training produces agents that engage in \emph{systematic, hypothesis-driven exploration}: testing individual actions, verifying constraints, and building comprehensive environment models. Task-oriented training produces agents that either terminate immediately (lacking a task reward signal) or execute shallow task-like routines that fail to discover environment structure. Second, the quality of the resulting environment knowledge directly determines downstream task performance: high-ECC exploration provides the grounded action semantics and object locations needed for efficient planning, while zero-ECC exploration leaves the agent to search blindly and perseverate on failed actions.


\clearpage

\end{document}